\documentclass[letterpaper]{article} 
\usepackage[draft]{aaai25}  
\usepackage{times}  
\usepackage{helvet}  
\usepackage{courier}  
\usepackage[hyphens]{url}  
\usepackage{graphicx} 
\usepackage{natbib}  
\usepackage{caption} 
\usepackage{algorithm}
\usepackage{algorithmic}
\usepackage{amsmath}
\usepackage{booktabs}
\usepackage{subcaption} 
\usepackage{pdfpages}

\usepackage{newfloat}
\usepackage{listings}
\DeclareCaptionStyle{ruled}{labelfont=normalfont,labelsep=colon,strut=off} 
\floatstyle{ruled}
\newfloat{listing}{tb}{lst}{}
\floatname{listing}{Listing}
\title{REPEAT: Improving Uncertainty Estimation in Representation Learning Explainability}
\author{
    Kristoffer K. Wickstr\o m\textsuperscript{\rm 1},
    Thea Brüsch\textsuperscript{\rm 2},
    Michael C. Kampffmeyer\textsuperscript{\rm 1, \rm 3},
    Robert Jenssen\textsuperscript{\rm 1, \rm 3, \rm 4}
}
\affiliations{
    \textsuperscript{\rm 1}Department of Physics and Technology, UiT  The Arctic University of Norway \\
    \textsuperscript{\rm 2}Department of Applied Mathematics and Computer Science, Technical University of Denmark \\
    \textsuperscript{\rm 3}Norwegian Computing Center, Oslo, Norway \\
    \textsuperscript{\rm 4}Pioneer Centre for AI, University of Copenhagen, Denmark
}

\newcommand{\etal}{\textit{et al.}}
\newcommand{\ul}[1]{\underline{#1}}

\begin{document}

\maketitle

\begin{abstract}
    Incorporating uncertainty is crucial to provide trustworthy explanations of deep learning models. Recent works have demonstrated how uncertainty modeling can be particularly important in the unsupervised field of representation learning explainable artificial intelligence (R-XAI). Current R-XAI methods provide uncertainty by measuring variability in the importance score. However, they fail to provide meaningful estimates of whether a pixel is certainly important or not. In this work, we propose a new R-XAI method called REPEAT that addresses the key question of whether or not a pixel is \textit{certainly} important. REPEAT leverages the stochasticity of current R-XAI methods to produce multiple estimates of importance, thus considering each pixel in an image as a Bernoulli random variable that is either important or unimportant. From these Bernoulli random variables we can directly estimate the importance of a pixel and its associated certainty, thus enabling users to determine certainty in pixel importance. Our extensive evaluation shows that REPEAT gives certainty estimates that are more intuitive, better at detecting out-of-distribution data, and more concise. Code is available at \url{https://github.com/Wickstrom/REPEAT}.
\end{abstract}

\section{Introduction}

Representation learning through self-supervision is the cornerstone of recent improvements in the computer vision domain~\cite{DeepCluster,MAE,Assran2023,Bardes2022}. Transforming images into a new representation has been shown to improve performance in a wide range of unsupervised tasks~\cite{Trosten2023CVPR, cmig-wickstrom, sehwag2021ssd}. Despite the benefits, unsupervised representation learning also suffers from some significant drawbacks, particularly a lack of explainability. Current methods in explainable artificial intelligence (XAI) are designed with supervised learning in mind, where a scalar model output is explained in relation to its input \cite{Petsiuk2018rise,Bach2015,integratedgradients}. Using these methods to explain \textit{representations} is either not possible or requires major modifications of the underlying algorithms~\cite{lfxai}.

Tackling this drawback has lead to a new direction within XAI, namely representation learning XAI (R-XAI). Methods within R-XAI solve the problem of explaining representations by either making adaptations of existing XAI methods~\cite{lfxai} or by designing new methods that are particularly designed to tackle the representation learning setting~\cite{relax, lin2023contrastive, enhanceREPS, moller2024finding}.

A key ingredient in recent R-XAI research is uncertainty estimation~\cite{relax}, where importance is accompanied by a corresponding uncertainty estimate. Providing an indication of certainty is highly desirable, for instance in safety critical areas such as healthcare \cite{tonekamboni, Kompa2021}. However, existing frameworks are limited to measuring the variation in the importance scores. This only gives an indication of how the numerical importance scores spread out, not if we are certain of importance. A more critical aspect is \textit{how certain are we that a pixel is important}. Consider an estimated importance map, where all pixels with importance scores higher than $2$ are considered important. Now, take one pixel with importance value $5.6 \pm 0.1$ and another with importance value $5.6 \pm 1.2$. 
Due to the higher variance of the second pixel, current R-XAI methods would assign high uncertainty to this pixel. 
However, since all values within the 95\% confidence interval of the pixel would still be above the importance threshold. As such, we would still be certain that this pixel is important, despite the higher uncertainty of the exact value.

\begin{figure*}[ht]
    \centering
    \includegraphics[width=\textwidth]{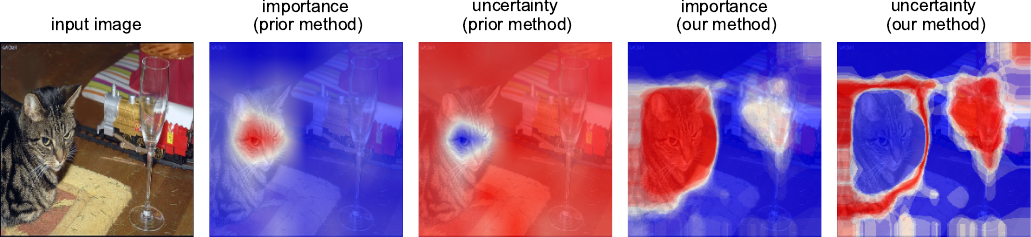}
    \caption{Motivating example to show the difference between current uncertainty estimation techniques in R-XAI and our proposed approach. An image from Pascal-VOC~\cite{Everingham2009} is encoded into a new representation using a ResNet18 feature extractor~\cite{resnet} and important pixels are determined. The importance (red indicates high and blue indicates low importance) is accompanied by uncertainties that specify how confident the importance of a pixel is. The results show that while the importance estimates somewhat agree, the uncertainty estimates are very different, which is due to the different type of uncertainty being captured (variation in pixel importance vs. certainty of importance).}
    \label{fig:motivation-example}
\end{figure*}

Fig. \ref{fig:motivation-example} shows an example that illustrates the distinction between these two questions, where a prior methods is compared to our proposed solution. The example shows which pixels are important for the representation of this image and corresponding uncertainty estimates that indicate how certain the importance is. Both methods mostly agree on the important input pixels, but have vastly different estimates of uncertainty. This shows the effect of modeling certainty in importance, as opposed to variability in importance scores.

\begin{figure*}[ht]
    \includegraphics[width=\linewidth]{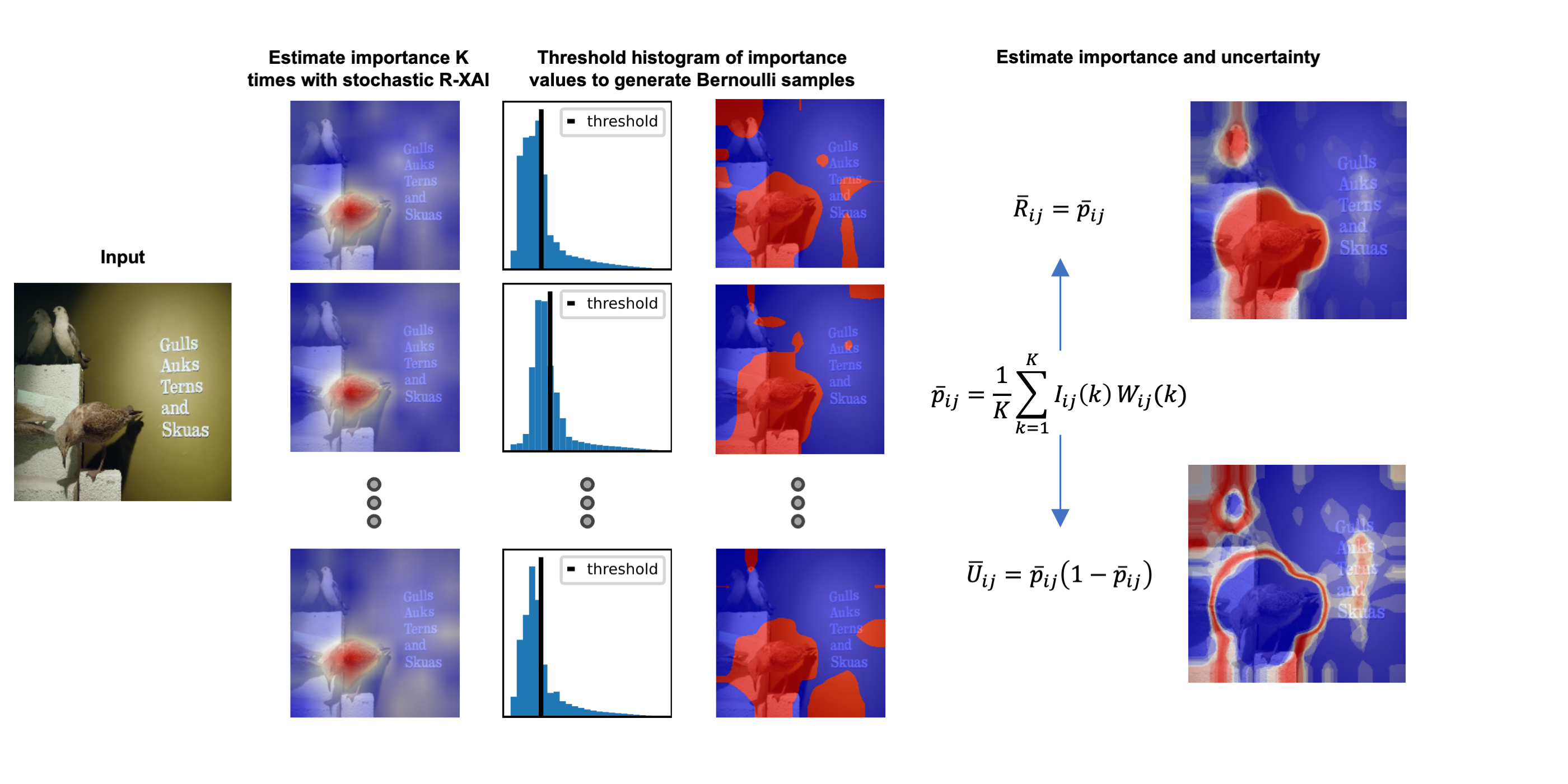}
    \caption{Overview of REPEAT. An image is transformed into a new representation and pixel importance is estimated using stochastic R-XAI. A histogram is constructed from pixel importance and Bernoulli samples are generated by thresholding the importance scores into foreground and background. Then, importance and uncertainty is estimated using the Bernoulli samples.}
    \label{fig:overview}
\end{figure*}

With the aim of answering the question of whether we are certain that a pixel is important or not, we present a new R-XAI method called REPEAT. The key idea of REPEAT is to consider each input pixel as a Bernoulli random variable (RV) that indicates if the pixel is important for the representation of the input image. To generate samples of these Bernoulli RVs, we leverage the stochasticity of prior R-XAI methods combined with classic image thresholding techniques to threshold an image into important and non-important pixels. By repeating the thresholding process on numerous importance estimates, a set of Bernoulli samples can be generated to estimate the probability of a pixel being important to the representation of an image and a corresponding uncertainty estimate which indicates how certain we are of the pixel being important or not. Fig. \ref{fig:overview} shows an overview of the proposed REPEAT framework. Our contributions are:

\begin{enumerate}
    \item A new R-XAI framework called REPEAT that models the importance of a pixel for the representation of an image as a Bernoulli RV. This RV indicates the probability of the pixel being important or not and indicates the certainty of the pixel being important.
    \item Extensive evaluation across numerous feature extractors and datasets and comparison with state-of-the-art baselines. Results show that REPEAT produces more intuitive uncertainty estimates that are better at detecting out-of-distribution data and has lower complexity, compared to other state-of-the-art methods.
    \item Evaluation on a downstream task where uncertainty is used to detect poisoned data in the unsupervised representation learning setting~\cite{he2023indiscriminate}.
\end{enumerate}

\section{Related Work}
\label{sec:related-work}

\textbf{Existing R-XAI literature:} There are two main approaches to extending the field of XAI to handle representations of data; adapt existing methods to handle the representation learning setting or design new methods designed for this particular use case. For adaptation approaches, \citeauthor{lfxai} proposed Label-Free Feature Importance, where an auxiliary scalar function allows standard XAI-methods to be used on each component of the representation~\cite{lfxai}. For R-XAI specific methods, \citeauthor{relax} introduced RELAX, where a representation is explained through similarity measurements between masked and unmasked representations of a particular image~\cite{relax}. \citeauthor{lin2023contrastive} extended existing methods in R-XAI to allow for explanation of a reference corpus in relation to a contrastive foil set~\cite{lin2023contrastive}. \citeauthor{moller2024finding} proposed a trainable explanations network aimed at increating the latency of the explanation process~\cite{moller2024finding}. \citeauthor{DeTomaso2021} proposed Hotspot, which is focused on explaining representations in single-cell genomics~\cite{DeTomaso2021}. Lastly, \citeauthor{enhanceREPS} introduced an aggregation method that generalizes attribution maps between any two convolutional layers of a neural network~\cite{enhanceREPS}. R-XAI has also been used in many applications, for instance in healthcare~\cite{chen2023interpreting, cmig-wickstrom, Weinberger2023} and business~\cite{Feng2023}.

\textbf{Uncertainty in XAI:} Several approaches have been proposed for modeling uncertainty in XAI. A number of works have investigated how to model uncertainty in surrogate-based XAI~\cite{LimeUc, posthocreliable, Wang2021uc, Schulz2022}, but these approaches are not transferable to the unsupervised setting. Other works have used Monte Carlo Dropout~\cite{mcdro} to estimate uncertainty in importance~\cite{Wickstrm2020} but this is restrictive because it requires Dropout~\cite{dropout} in the feature extractor. Another work has used ensembles~\cite{simpleensemble} for uncertainty estimation in XAI~\cite{9284514}, but this is not directly applicable in this context since we are interested in explaining a single feature extractor. In the general field of uncertainty modeling, test-time-augmentation has been shown to be a generally-applicable and effective tool for uncertainty estimation~\cite{ttadropout, kahl2024values}, but has not been explored in the context of R-XAI. The R-XAI framework RELAX~\cite{relax} provides uncertainty estimates with its importance scores, but the uncertainty estimates measure the variability in the importance scores and not certainty in pixel importance.

\section{REPEAT: a new method for R-XAI with Improved Uncertainty Estimation}

We present REPEAT, a new method for R-XAI that indicates which pixels in an image are most important for the representation of the image and provides uncertainty estimates that specify if a pixel is certainly important or not.

\subsection{Interpreting input pixels as Bernoulli RVs}

Let $X_{ij}$ be a RV following a Bernoulli distribution such that $\Pr(X_{ij} = 1)=p_{ij}$ indicates the probability of pixel $\{i, j\}$ being important for the representation $\mathbf{h}$ of $\mathbf{X}$ by the feature extractor $f$, and $\Pr(X_{ij} = 0)=q_{ij}=(1-p_{ij})$ indicates the opposite case. We consider the importance of a pixel as:

\begin{equation}\label{eq:rtr-expectation}
    R_{ij} = \mathrm{E} \left[X_{ij} \right]=p_{ij}.
\end{equation}
Furthermore, we consider the uncertainty associated with the importance as:

\begin{equation}\label{eq:rtr-variance}
    U_{ij} = \mathrm{Var} \left[ X_{ij} \right]=p_{ij}(1-p_{ij}).
\end{equation}
The value of $p_{ij}$ is unknown, but can be estimated from data. To perform this estimation, we require realizations of $X_{ij}$. The following subsection presents how to generate these realizations.

\subsection{Generating Samples using 
Stochastic R-XAI and Thresholding}
\label{sec:generate-B}

To estimate $p_{ij}$ we require samples that indicate whether or not a pixel is important to the representation of an image. We propose to leverage a base stochastic R-XAI method to generate samples as follows:

\begin{equation}\label{eq:rtr-indicator}
I_{ij}\left(k\right)=
\begin{cases}
1 \hspace{0.25cm} \text{if} \hspace{0.25cm} \Bar{R}_{ij}^{\text{base}}(k) \geq \tau \\
0 \hspace{0.25cm} \text{else}
\end{cases}. 
\end{equation}
Here, for the $k^{\text{th}}$ realization, $I_{ij}\left(k\right)$ is an indicator function that activates if the importance score is above a certain threshold $\tau$ (see next subsection for threshold selection), and $\Bar{R}_{ij}^{\text{base}}(k)$ is the estimated importance score from the base stochastic R-XAI method. If the scores are above the threshold, the pixel is considered important for the representation of the image in question. The stochasticity requirement for the base R-XAI method is critical, since this will ensure diversity and allow new realizations of the RV $X_{ij}$ to be generated. By repeating this process we obtain a set of samples that can be used to estimate $p_{ij}$. The intuition is that pixels which are assigned high importance across numerous realizations will have a high probability of being important. Similarly, pixels that are regularly assigned low scores will have a low probability of being important. And importantly, pixels that fluctuate above and below the threshold will be highlighted as having high uncertainty.

\subsection{Setting the Threshold}

\begin{figure}[htb!]
    \centering
    \includegraphics[width=0.975\linewidth]{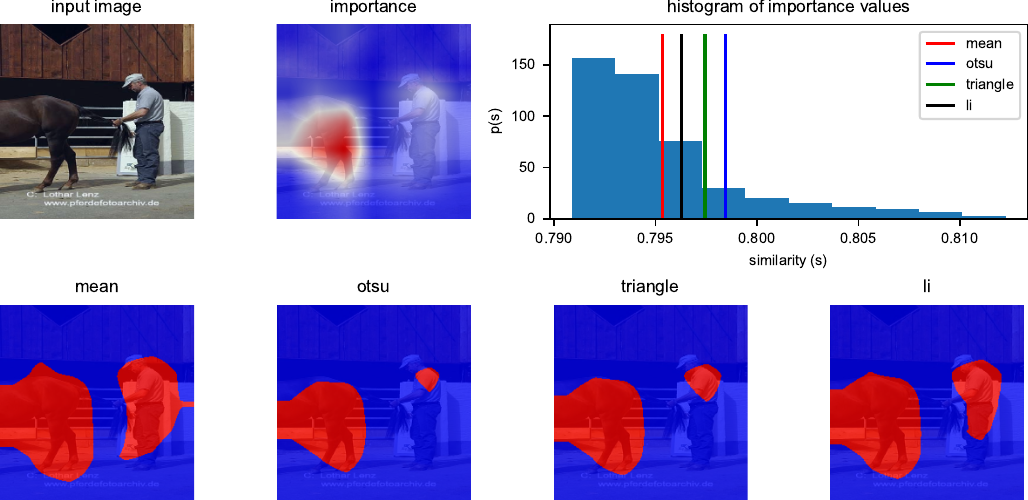}
    \caption{Top row: input image from Pascal VOC~\cite{Everingham2009}, importance and histogram of similarities. Bottom row: thresholding of importance.}
    \label{fig:threshold-example}
\end{figure}

Finding a proper value for $\tau$ in Eq. \ref{eq:rtr-indicator} is crucial to generate meaningful samples. We propose to approach this from a foreground-background thresholding perspective, where we consider important pixels as foreground pixels and unimportant pixels as background pixels. Thresholding is a classic problem in image processing with a vast amount of established literature~\cite{dip-book}. Thresholding algorithms can roughly be separated into two categories; histogram-based or local methods. Histogram-based methods use the histogram of pixel intensities, while local methods process each pixel by considering its neighborhood. Local methods are more computationally demanding, and since we will repeat the thresholding procedure multiple times, we will only focus on histogram-based methods. We consider four widely used approaches to histogram-based image thresholding: mean~\cite{Glasbey1993}, Otsu's method~\cite{otsu}, triangle method~\cite{Zack1977}, and Li's method~\cite{Li1993, Li1998}. In App. A we provide a more detailed explanation of these methods. Fig. \ref{fig:threshold-example} displays an example of the thresholding procedure. Here, we feed an image into a ResNet50 encoder~\cite{resnet}, and determine importance using an existing stochastic R-XAI method~\cite{relax}. We compute a histogram based on the importance values of all pixels and apply the four thresholding methods to set a threshold for separating foreground (important pixels) from background (unimportant pixels). In this example, the mean thresholding is more conservative and keeps more pixels as important, while Otsu's method produces a higher threshold value that assigns more pixels as unimportant. We found that this behavior was quite consistent across several R-XAI methods and datasets. In App. B, we compared all four methods and their potential in REPEAT, and found that mean thresholding yielded the best performance. Therefore, we use mean thresholding as the standard thresholding method for the remainder of this work.

\subsection{Importance and Uncertainty With REPEAT}

With repeated use of Eq. \ref{eq:rtr-indicator}, new data points can be generated to estimate $p_{ij}$. Assuming $K$ importance maps are generated, we propose to use a weighted sample mean that takes into consideration the base importance scores as follows:

\begin{equation}\label{eq:rtr-explanation}
    \Bar{p}_{ij} = \frac{1}{K}\sum\limits_{k=1}^K I_{ij}(k) W_{ij}(k) = \Bar{R}_{ij},
\end{equation}
where

\begin{equation}\label{eq:rtr-weight}
    W_{ij}(k) = \frac{\Bar{R}_{ij}^{\text{base}}(k)}{A(k)}
\end{equation}
and $A(k)$ is the maximum value of $\Bar{\mathbf{R}}^{\text{base}}(k)$. Weighting the indicator function with the importance scores facilitates ranking among the most important pixels in an image, and scaling by the maximum value ensures that the importance scores are comparable across the repeated explanations. In App. A, we also provide an interpretation of REPEAT from the perspective of multiple kernel learning~\cite{mkl}, which shows that REPEAT can be understood as a weighted linear scoring function in a reproducing kernel Hilbert space. Finally, the uncertainty of pixel importance in REPEAT can be calculated in the standard way for Bernoulli RVs:

\begin{equation}\label{eq:rtr-uncertainty}
    \Bar{U}_{ij}^{\text{rtr}} = \Bar{p}_{ij}(1-\Bar{p}_{ij}).
\end{equation}

\section{Evaluation Protocol and Experimental Setup}\label{sec:evaluation}

Here, we describe how we evaluate and compare REPEAT to state-of-the-art alternatives and detail our experimental setup.

\begin{table*}[htb!]
\parbox{.475\linewidth}{
\centering
\begin{tabular}{lcccc}
\toprule
 explainability method & \multicolumn{1}{c}{resnet50} & & \multicolumn{1}{c}{vit} \\ \midrule
saliency* & -0.16 &  & -0.10 \\
guided backpropagation* & -0.23 &  & -0.10  \\
integrated gradients* & -0.17 &  & -0.19  \\
RELAX & \underline{-0.08} &  & -0.07 \\
REPEAT & \textbf{0.48} &  & \textbf{0.13} \\
\bottomrule
\end{tabular}%
}
\hfill
\parbox{.475\linewidth}{
\centering
\begin{tabular}{lcccc}
\toprule
 explainability method & \multicolumn{1}{c}{resnet50} & & \multicolumn{1}{c}{vit} \\ \midrule
saliency* & -0.16 &  & -0.11 \\
guided backpropagation* & -0.21 &  & -0.11 \\
integrated gradients* & -0.18 &  & -0.19 \\
RELAX & -0.08 &  & -0.08 \\
REPEAT & \textbf{0.44} &  & \textbf{0.13} \\
\bottomrule
\end{tabular}%
}
\vspace{0.1cm}
\caption{Results for R-XAI method for the sanity check (higher is better) of uncertainty estimates on PASCAL-VOC (left) and MS-COCO (right) with a ResNet50 and a ViT encoder. The best and second best performance for each column are indicated by \textbf{bold} and \ul{underlined}, respectively. The * highlights that these methods are adapted to the R-XAI setting using Label-Free Feature Importance~\cite{lfxai}.}
\label{tab:sanity}
\end{table*}

\paragraph{Evaluation Protocol}

We describe how we evaluate and compare uncertainty estimates. Our evaluation follows well known tasks in both the uncertainty and the XAI literature.

\textbf{Sanity check:} The Model Parameter Randomisation Test (MPRT)~\cite{mprt} is widely used in XAI to investigate if XAI method behaves as expected~\cite{barkan2023iccv,NEURIPS2023c2eac51b,sanityref}. The general idea is to see if explanations deteriorate when the parameters of a model are randomized before the decision of the model is explained. However, the MPRT can be highly computationally demanding~\cite{hedstroem2023sanity}, which makes it difficult to provide a comprehensive analysis. Therefore, we instead use the recently proposed efficient MPRT (eMPRT)~\cite{hedstroem2023sanity}. The eMPRT compares the relative rise in explanation complexity for an explanation of a trained model compared to a completely randomized model. The intuition is that explanations of random models should be mostly random and therefore have high entropy, while explanations of a trained model should be more focused and therefore have lower entropy. A high positive value is desirable, as it indicates that the explanations of the random model are more complex, while a negative value indicates that the explanations of the trained model are more complex and is not desirable. Originally, both the MPRT and the eMPRT were designed for evaluation of explanations, but here we use eMPRT on the uncertainty estimates.

\textbf{Out-of-distribution detection:} A standard task in uncertainty estimation is out-of-distribution (OOD) detection~\cite{simpleensemble, swag}. When presented with unfamiliar data this should be reflected in the uncertainty. We follow the approach of prior works~\cite{simpleensemble, hein} and measure to what degree the uncertainty estimates can be used to differentiate between importance scores for in-distribution and OOD data, based on the uncertainty estimates. Since we are in the unsupervised setting, we propose to detect the OOD data using a Gaussian mixture model~\cite{Reynolds2009} with two components and treat the component with the highest mean as the OOD detector. We choose the component with the highest mean, since we expect the OOD data to have the highest uncertainty.

\textbf{Complexity:} In the greater XAI literature, a desirable property of explanations is conciseness, often referred to as low complexity~\cite{complex1, complexentropy}. The same property is also desirable for uncertainty estimates, since a model that is uncertain about all importance scores is not informative. Instead, the model should be confident about clearly important and unimportant pixels, and only uncertain about some critical pixels where there is ambiguity. In this work, we measure complexity following the standard approach proposed by Bhatt \etal~\cite{complexentropy}, where complexity is calculated by taking the entropy of the uncertainties for an image.

\begin{figure*}[htb!]
 \centering
 \begin{subfigure}{0.475\textwidth}
     \centering
     \includegraphics[width=0.975\columnwidth]{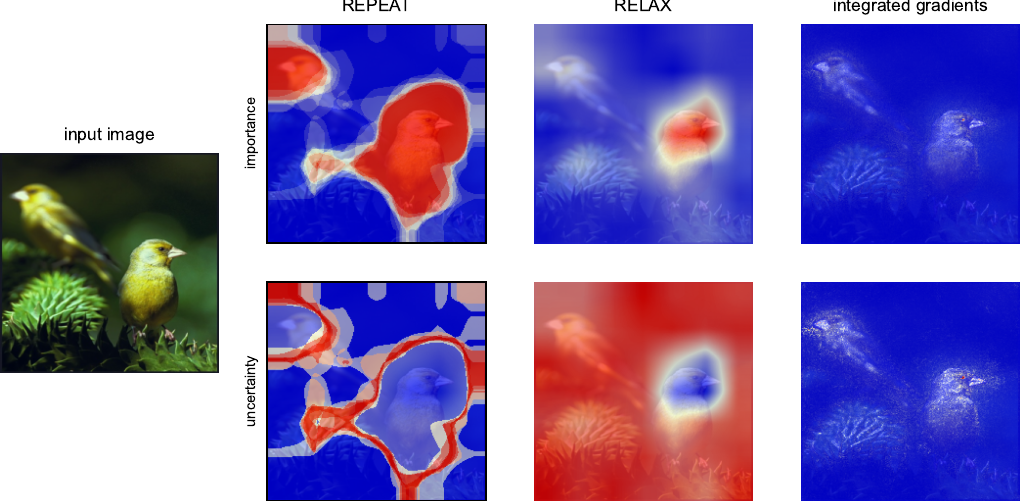}
     \caption{}
     \label{fig:qualitative1}
 \end{subfigure}
 \hfill
 \begin{subfigure}{0.475\textwidth}
     \centering
     \includegraphics[width=0.975\columnwidth]{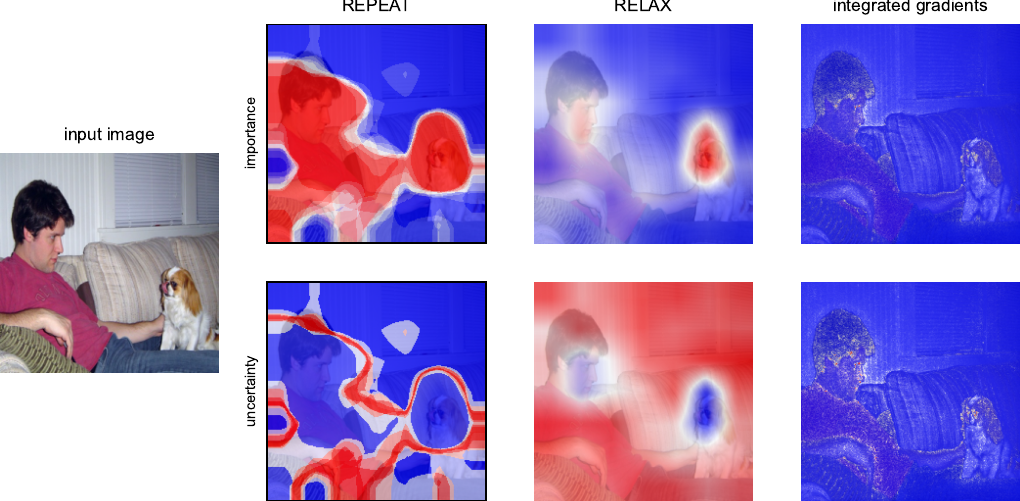}
     \caption{}
     \label{fig:qualitative2}
 \end{subfigure}
    \caption{Qualitative examples on images from PASCAL VOC.}
    \label{fig:qualitative}
\end{figure*}

\begin{table*}[htb!]
\parbox{.475\linewidth}{
\centering
\begin{tabular}{lcccc}
\toprule
 explainability method & \multicolumn{1}{c}{resnet50} & & \multicolumn{1}{c}{vit} \\ \midrule
saliency* & 0.797 &  & \underline{0.002} \\
guided backpropagation* & 0.782 &  & 0.001  \\
integrated gradients* & \underline{0.998} &  & \underline{0.002} \\
RELAX & 0.000 &  & 0.000 \\
REPEAT & \textbf{1.000} &  & \textbf{0.973} \\
\bottomrule
\end{tabular}%
}
\hfill
\parbox{.475\linewidth}{
\centering
\begin{tabular}{lcccc}
\toprule
 explainability method & \multicolumn{1}{c}{resnet50} & & \multicolumn{1}{c}{vit} \\ \midrule
saliency* & 0.969 &  & 0.001 \\
guided backpropagation* & 0.823 &  & 0.000 \\
integrated gradients* & \underline{0.999} &  & \underline{0.002} \\
RELAX & 0.000 &  & 0.000 \\
REPEAT & \textbf{1.000} &  & \textbf{0.939} \\
\bottomrule
\end{tabular}%
}
\vspace{0.1cm}
\caption{Results for R-XAI methods for OOD detection using EuroSAT as the OOD dataset. The table shows AUROC when classifying in-domain (PASCAL-VOC (left) or MS-COCO (right)) vs out-of-domain clusters using a Gaussian mixture model (higher is better). The best and second best performance for each column are indicated by \textbf{bold} and \ul{underlined}, respectively. The * highlights that these methods are adapted to the R-XAI setting using Label-Free Feature Importance~\cite{lfxai}.}
\label{tab:ood-results}
\end{table*}

\begin{figure*}[htb!]
    \centering
    \includegraphics[width=\textwidth]{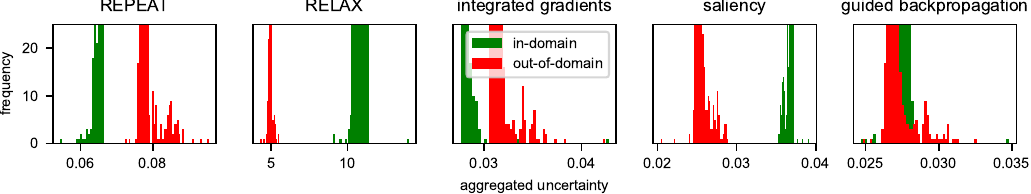}
    \caption{Histogram of aggregated uncertainty scores for in-distribution (PASCAL-VOC in green) and OOD (EuroSAT in red) data. This example illustrates how REPEAT gives the clearest separation between in-distribution and OOD data under the assumption that OOD data should have highest uncertainty.}
    \label{fig:ood}
\end{figure*}

\paragraph{Experimental Setup} We investigate the performance of REPEAT and competing methods across numerous feature extractors, datasets, and baselines that are described below.

\textbf{REPEAT design choices:} In all presented results, we generate K=10 realizations of the Bernoulli RVs and use the mean to perform the thresholding. Both of these choices are determined by quantitative evaluation that is reported in App. B. As the base stochastic R-XAI method we use RELAX~\cite{relax}, due to its high performance in recent works. However, to demonstrate REPEAT's ability to leverage any stochastic R-XAI method, we also evaluated the performance of REPEAT with Kernel SHAP~\cite{kernelshap}, which we show in the results. We also reduce the computational demand of RELAX by developing a new bound on the estimation error in RELAX (see App. A).

\textbf{Datasets:} We use four widely used computer vision datasets; MS-COCO~\cite{Lin2014}, Pascal-VOC~\cite{Everingham2009}, EuroSAT~\cite{helber2018introducing}, and FashionMNIST~\cite{fashionmnist}.

\textbf{Baseline XAI methods:} We compare the performance of REPEAT with several strong baselines. First, RELAX~\cite{relax}, which is designed for the R-XAI setting. Apart from RELAX, all other methods are adopted to the R-XAI setting using Label-Free Feature Importance~\cite{lfxai}. These baselines are: Saliency~\cite{kaihansen95}, Guided-Backpropagation~\cite{guidebackprop}, and Integrated Gradients~\cite{integratedgradients}.

\textbf{Feature extractors:} We leverage two state-of-the-art feature extractors to create the representations we want to explain; the ResNet50~\cite{resnet} and the Vision Transformer (ViT)~\cite{dosovitskiy2021an}. For the ResNet50, we take the representation to be the output of the adaptive pooling layer at the end of convolutional neural network backbone. For the ViT, we use the base model and treat the classification token as the representation. For simplicity and reproducibility, we use the pretrained weights from Pytorch~\cite{pytorch} for supervised classification of ImageNet~\cite{deng2009imagenet}.

\textbf{Baseline uncertainty estimation:} REPEAT and RELAX provide uncertainty estimates as part of their framework. The remaining baseline methods do not have this capability, and external methods must be used to estimate uncertainty in their importance scores. Due to its flexibility and performance~\cite{kahl2024values}, we propose to use test-time augmentation~\cite{Abdar2021} to estimate uncertainty for the remaining baseline methods. Specifically, we follow Wang \etal~\cite{ttadropout}, where Dropout is applied to the input (Dropout probability of 0.5). Here, we create 10 Dropout-versions of each image and calculate importance using the baseline methods. Uncertainty is computed by taking the standard deviation across all 10 importance maps.

\section{Results}
\label{sec:results}

This section present the main results of our work. First, we present the results of the outlined evaluation protocol. Then, we show that REPEAT is also applicable beyond RELAX. In all experiments, we randomly sample 1000 images from the dataset used for evaluation. We found that this was enough samples to provide reliable estimates of performance while still being computationally tractable. Due to their inherent stochasticity, RELAX and REPEAT experiments were repeated 3 times. This revealed (shown in App. B) that performance was stable with a standard deviation less than the decimal precision reported here. Therefore, for clarity we do not report the standard deviation here. Also, in App. B we evaluate the performance of the explanations produced by REPEAT, which shows good performance.

\begin{table*}[htb]
\parbox{.475\linewidth}{
\centering
\begin{tabular}{lcccc}
\toprule
 explainability method & \multicolumn{1}{c}{resnet50} & & \multicolumn{1}{c}{vit} \\ \midrule
saliency* &10.52 &  & \underline{10.46}  \\
guided backpropagation* &\underline{10.41} &  & \underline{10.46}  \\
integrated gradients* &10.44  &  & 10.49  \\
RELAX &10.82 &  & 10.82   \\
REPEAT &\textbf{9.97}  &  & \textbf{9.99}   \\
\bottomrule
\end{tabular}%
}
\hfill
\parbox{.475\linewidth}{
\centering
\begin{tabular}{lcccc}
\toprule
 explainability method & \multicolumn{1}{c}{resnet50} & & \multicolumn{1}{c}{vit} \\ \midrule
saliency* &10.45 &  & \underline{10.40}  \\
guided backpropagation* &\underline{10.35} &  & \underline{10.40} \\
integrated gradients* &10.37 &  & 10.42   \\
RELAX &10.75  &  & 10.74   \\
REPEAT &\textbf{9.91}  &  & \textbf{9.94} \\
\bottomrule
\end{tabular}%
}
\vspace{0.1cm}
\caption{Results for R-XAI method for complexity (lower is better) of uncertainty estimates on PASCAL-VOC (left) and MS-COCO (right) with a ResNet50 and a ViT encoder. The best and second best performance for each column are indicated by \textbf{bold} and \ul{underlined}, respectively. The * highlights that these methods are adapted to the R-XAI setting using Label-Free Feature Importance~\cite{lfxai}.}
\label{tab:complexity}
\end{table*}

\begin{table*}[]
\parbox{.475\linewidth}{
\centering
\begin{tabular}{lcccc}
\toprule
 explainability method & \multicolumn{1}{c}{resnet50} & & \multicolumn{1}{c}{vit} \\ \midrule
REPEAT (Kernel SHAP) & \underline{0.999} &  & \textbf{0.999}   \\
REPEAT (RELAX) & \textbf{1.000}  &  & \underline{0.973}   \\
\bottomrule
\end{tabular}%
}
\hfill
\parbox{.475\linewidth}{
\centering
\begin{tabular}{lcccc}
\toprule
 explainability method & \multicolumn{1}{c}{resnet50} & & \multicolumn{1}{c}{vit} \\ \midrule
REPEAT (Kernel SHAP) & \textbf{1.000} &  & \textbf{1.000} \\
REPEAT (RELAX) & \textbf{1.000} &  & \underline{0.939} \\
\bottomrule
\end{tabular}%
}
\vspace{0.1cm}
\caption{Results for R-XAI methods for OOD detection using the EuroSAT dataset as the OOD dataset, with different base R-XAI methods in REPEAT. The table shows AUROC when classifying in-domain (VOC or COCO) vs out-of-domain clusters using a Gaussian mixture model. The best and second best performance for each column are indicated by \textbf{bold} and \ul{underlined}, respectively.}
\label{tab:kernel-shap}
\end{table*}

\begin{table*}[hbt!]
\parbox{.475\linewidth}{
\centering
\begin{tabular}{lcccc}
\toprule
 explainability method & \multicolumn{1}{c}{resnet50} & & \multicolumn{1}{c}{vit} \\ \midrule
REPEAT (Kernel SHAP) & \underline{10.40} &  & \underline{10.36}   \\
REPEAT (RELAX) &\textbf{9.97}  &  & \textbf{9.99}   \\
\bottomrule
\end{tabular}%
}
\hfill
\parbox{.475\linewidth}{
\centering
\begin{tabular}{lcccc}
\toprule
 explainability method & \multicolumn{1}{c}{resnet50} & & \multicolumn{1}{c}{vit} \\ \midrule
REPEAT (Kernel SHAP) & \underline{10.28} &  & \underline{10.28}  \\
REPEAT (RELAX) &\textbf{9.91}  &  & \textbf{9.94} \\
\bottomrule
\end{tabular}%
}
\vspace{0.1cm}
\caption{Results for R-XAI method for complexity (lower is better) of uncertainty estimates for different base R-XAI methods in REPEAT on PASCAL-VOC (left) and MS-COCO (right) with a ResNet50 and a ViT encoder. The best and second best performance for each column are indicated by \textbf{bold} and \ul{underlined}, respectively.}
\label{tab:ks-complexity}
\end{table*}

\subsection{Quantitative Evaluation}

\textbf{Sanity check:} Tab. \ref{tab:sanity} shows the results for the sanity check. The results show that REPEAT outperforms all baselines methods across all datasets and encoders. Interestingly, REPEAT is the only method that provides uncertainty estimates that pass the sanity check. This highlights that the proposed method brings a significant advantage compared to competing methods. Lastly, Fig. \ref{fig:qualitative} shows some qualitative examples, and more can be found in App. B. 

\textbf{OOD detection:} Tab. \ref{tab:ood-results} shows the results of the OOD experiment with Eurosat as the OOD data. A similar experiment is shown in App. B with FashionMNIST as the OOD data. The results show how REPEAT clearly outperforms all other methods. Particularly RELAX has low performance, because it gives lower uncertainty scores to the OOD data, and thus falsely classifies in-domain data as OOD. Fig. \ref{fig:ood} displays an example of the distribution of aggregated uncertainties for all methods. Note how the in-distribution and OOD data are clearly separable, with the uncertainty for the OOD data being much larger than the in-distribution data. In contrast, both RELAX and Saliency has lower uncertainty for the OOD data, which is the complete opposite of the desired behavior. For Guided Backpropagation, the two distributions are indistinguishable. Integrated Gradients is adequate at separating the two distributions, but has much less separation compared to REPEAT. Motivated by REPEAT's successful OOD detection abilities, we also conducted experiments on poisoned data~\cite{poisonreivew} with encouraging performance. These results can be seen in App. B, and show that uncertainty is essential to obtain good performance on this downstream task.

\textbf{Complexity evaluation:} Tab. \ref{tab:complexity} displays the result of complexity evaluation of the different uncertainty estimates, which shows that REPEAT outperforms all method across all settings, thus providing more concise uncertainty estimates compared to existing methods.

\textbf{REPEAT beyond RELAX:} In this work, we have focused on RELAX as the base R-XAI method in REPEAT. However, REPEAT is more general and can be used with any stochastic R-XAI method. To illustrate this, we have conducted the same OOD detection and complexity experiments as earlier but with Kernel-SHAP~\cite{kernelshap} (adapted to the R-XAI setting using Label-Free Feature Importance~\cite{lfxai}) as the base stochastic R-XAI method. These results are shown in Tab. \ref{tab:kernel-shap} and Tab. \ref{tab:ks-complexity}, and demonstrate the flexibility of REPEAT. The performance for OOD detection is very similar for RELAX compared to Kernel-SHAP, but RELAX gives lower complexity compared to using Kernel-SHAP as the base R-XAI method.

\section{Conclusion}
Current methods for determining certainty in pixel importance are limited, as they only estimate the variability in the importance values. This reduces the reliability of R-XAI, as users cannot decide if a pixel is certainly important or not. In this work, we proposed a new method called REPEAT that addresses this limitation. REPEAT treats each pixel in an image as a Bernoulli RV that is either important or unimportant to the representation of the image. From these Bernoulli RV we can directly estimate the importance of a pixel and its associated certainty, thus enabling users to ascertain certainty in pixel importance. We conducted an extensive evaluation which showed that REPEAT provides more intuitive uncertainty estimates that are better at identifying OOD data and with lower complexity. Further, we also show that REPEAT works effectively with different types of stochastic R-XAI methods. We believe REPEAT can play an important role in moving the field of R-XAI forward.

\bibliography{aaai25}

\begin{thebibliography}{63}
\providecommand{\natexlab}[1]{#1}

\bibitem[{Abdar et~al.(2021)Abdar, Pourpanah, Hussain, Rezazadegan, Liu, Ghavamzadeh, Fieguth, Cao, Khosravi, Acharya, Makarenkov, and Nahavandi}]{Abdar2021}
Abdar, M.; Pourpanah, F.; Hussain, S.; Rezazadegan, D.; Liu, L.; Ghavamzadeh, M.; Fieguth, P.; Cao, X.; Khosravi, A.; Acharya, U.~R.; Makarenkov, V.; and Nahavandi, S. 2021.
\newblock A review of uncertainty quantification in deep learning: Techniques, applications and challenges.
\newblock \emph{Information Fusion}, 76: 243–297.

\bibitem[{Adebayo et~al.(2018)Adebayo, Gilmer, Muelly, Goodfellow, Hardt, and Kim}]{mprt}
Adebayo, J.; Gilmer, J.; Muelly, M.; Goodfellow, I.; Hardt, M.; and Kim, B. 2018.
\newblock Sanity checks for saliency maps.
\newblock In \emph{Proceedings of the 32nd International Conference on Neural Information Processing Systems}, NIPS'18, 9525–9536. Red Hook, NY, USA: Curran Associates Inc.

\bibitem[{Assran et~al.(2023)Assran, Duval, Misra, Bojanowski, Vincent, Rabbat, LeCun, and Ballas}]{Assran2023}
Assran, M.; Duval, Q.; Misra, I.; Bojanowski, P.; Vincent, P.; Rabbat, M.; LeCun, Y.; and Ballas, N. 2023.
\newblock Self-Supervised Learning from Images with a Joint-Embedding Predictive Architecture.
\newblock In \emph{2023 IEEE/CVF Conference on Computer Vision and Pattern Recognition (CVPR)}. IEEE.

\bibitem[{Bach et~al.(2015)Bach, Binder, Montavon, Klauschen, M\"{u}ller, and Samek}]{Bach2015}
Bach, S.; Binder, A.; Montavon, G.; Klauschen, F.; M\"{u}ller, K.-R.; and Samek, W. 2015.
\newblock On Pixel-Wise Explanations for Non-Linear Classifier Decisions by Layer-Wise Relevance Propagation.
\newblock \emph{PLOS ONE}, 10(7): e0130140.

\bibitem[{Bardes, Ponce, and LeCun(2022)}]{Bardes2022}
Bardes, A.; Ponce, J.; and LeCun, Y. 2022.
\newblock VICRegL: Self-Supervised Learning of Local Visual Features.
\newblock In Koyejo, S.; Mohamed, S.; Agarwal, A.; Belgrave, D.; Cho, K.; and Oh, A., eds., \emph{Advances in Neural Information Processing Systems}, 8799--8810. Curran Associates, Inc.

\bibitem[{Barkan et~al.(2023{\natexlab{a}})Barkan, Elisha, Asher, Eshel, and Koenigstein}]{barkan2023iccv}
Barkan, O.; Elisha, Y.; Asher, Y.; Eshel, A.; and Koenigstein, N. 2023{\natexlab{a}}.
\newblock Visual Explanations via Iterated Integrated Attributions.
\newblock In \emph{Proceedings of the IEEE/CVF International Conference on Computer Vision (ICCV)}, 2073--2084.

\bibitem[{Barkan et~al.(2023{\natexlab{b}})Barkan, Elisha, Weill, Asher, Eshel, and Koenigstein}]{sanityref}
Barkan, O.; Elisha, Y.; Weill, J.; Asher, Y.; Eshel, A.; and Koenigstein, N. 2023{\natexlab{b}}.
\newblock Deep Integrated Explanations.
\newblock \emph{Proceedings of the 32nd ACM International Conference on Information and Knowledge Management}.

\bibitem[{Bertolini, Clevert, and Montanari(2023)}]{enhanceREPS}
Bertolini, M.; Clevert, D.-A.; and Montanari, F. 2023.
\newblock Explaining, Evaluating and Enhancing Neural Networks’ Learned Representations.
\newblock In \emph{Artificial Neural Networks and Machine Learning – ICANN 2023: 32nd International Conference on Artificial Neural Networks, Heraklion, Crete, Greece, September 26–29, 2023, Proceedings, Part V}, 269–287. Berlin, Heidelberg: Springer-Verlag.
\newblock ISBN 978-3-031-44191-2.

\bibitem[{Bhatt, Weller, and Moura(2021)}]{complexentropy}
Bhatt, U.; Weller, A.; and Moura, J. M.~F. 2021.
\newblock Evaluating and aggregating feature-based model explanations.
\newblock In \emph{Proceedings of the Twenty-Ninth International Joint Conference on Artificial Intelligence}, IJCAI'20.
\newblock ISBN 9780999241165.

\bibitem[{Caron et~al.(2018)Caron, Bojanowski, Joulin, and Douze}]{DeepCluster}
Caron, M.; Bojanowski, P.; Joulin, A.; and Douze, M. 2018.
\newblock Deep Clustering for Unsupervised Learning of Visual Features.
\newblock In \emph{Proceedings of the European Conference on Computer Vision (ECCV)}.

\bibitem[{Chalasani et~al.(2020)Chalasani, Chen, Chowdhury, Wu, and Jha}]{complex1}
Chalasani, P.; Chen, J.; Chowdhury, A.~R.; Wu, X.; and Jha, S. 2020.
\newblock Concise Explanations of Neural Networks using Adversarial Training.
\newblock In III, H.~D.; and Singh, A., eds., \emph{Proceedings of the 37th International Conference on Machine Learning}, volume 119 of \emph{Proceedings of Machine Learning Research}, 1383--1391. PMLR.

\bibitem[{Chen et~al.(2023)Chen, Bijlani, Kouchaki, and Barnaghi}]{chen2023interpreting}
Chen, Y.; Bijlani, N.; Kouchaki, S.; and Barnaghi, P. 2023.
\newblock Interpreting Differentiable Latent States for Healthcare Time-series Data.
\newblock In \emph{ICML 3rd Workshop on Interpretable Machine Learning in Healthcare (IMLH)}.

\bibitem[{Crabb{\'e} and van~der Schaar(2022)}]{lfxai}
Crabb{\'e}, J.; and van~der Schaar, M. 2022.
\newblock Label-Free Explainability for Unsupervised Models.
\newblock In \emph{Proceedings of the 39th International Conference on Machine Learning}, Proceedings of Machine Learning Research, 4391--4420. PMLR.

\bibitem[{Deng et~al.(2009)Deng, Dong, Socher, Li, Li, and Fei-Fei}]{deng2009imagenet}
Deng, J.; Dong, W.; Socher, R.; Li, L.-J.; Li, K.; and Fei-Fei, L. 2009.
\newblock Imagenet: A large-scale hierarchical image database.
\newblock In \emph{IEEE Computer Vision and Pattern Recognition}, 248--255. Ieee.

\bibitem[{DeTomaso and Yosef(2021)}]{DeTomaso2021}
DeTomaso, D.; and Yosef, N. 2021.
\newblock Hotspot identifies informative gene modules across modalities of single-cell genomics.
\newblock \emph{Cell Systems}, 12: 446--456.e9.

\bibitem[{Dosovitskiy et~al.(2021)Dosovitskiy, Beyer, Kolesnikov, Weissenborn, Zhai, Unterthiner, Dehghani, Minderer, Heigold, Gelly, Uszkoreit, and Houlsby}]{dosovitskiy2021an}
Dosovitskiy, A.; Beyer, L.; Kolesnikov, A.; Weissenborn, D.; Zhai, X.; Unterthiner, T.; Dehghani, M.; Minderer, M.; Heigold, G.; Gelly, S.; Uszkoreit, J.; and Houlsby, N. 2021.
\newblock An Image is Worth 16x16 Words: Transformers for Image Recognition at Scale.
\newblock In \emph{International Conference on Learning Representations}.

\bibitem[{Everingham et~al.(2009)Everingham, Gool, Williams, Winn, and Zisserman}]{Everingham2009}
Everingham, M.; Gool, L.~V.; Williams, C. K.~I.; Winn, J.; and Zisserman, A. 2009.
\newblock The Pascal Visual Object Classes ({VOC}) Challenge.
\newblock \emph{International Journal of Computer Vision}, 303--338.

\bibitem[{Feng, Li, and Zhang(2023)}]{Feng2023}
Feng, X.~F.; Li, C.; and Zhang, S. 2023.
\newblock Visual Uniqueness in Peer-to-Peer Marketplaces: Machine Learning Model Development, Validation, and Application.
\newblock \emph{SSRN Electronic Journal}.

\bibitem[{Gal and Ghahramani(2016)}]{mcdro}
Gal, Y.; and Ghahramani, Z. 2016.
\newblock Dropout as a Bayesian Approximation: Representing Model Uncertainty in Deep Learning.
\newblock In Balcan, M.~F.; and Weinberger, K.~Q., eds., \emph{Proceedings of The 33rd International Conference on Machine Learning}, Proceedings of Machine Learning Research, 1050--1059. New York, New York, USA: PMLR.

\bibitem[{Glasbey(1993)}]{Glasbey1993}
Glasbey, C. 1993.
\newblock An Analysis of Histogram-Based Thresholding Algorithms.
\newblock \emph{CVGIP: Graphical Models and Image Processing}, 55(6): 532–537.

\bibitem[{Goldblum et~al.(2023)Goldblum, Tsipras, Xie, Chen, Schwarzschild, Song, Madry, Li, and Goldstein}]{poisonreivew}
Goldblum, M.; Tsipras, D.; Xie, C.; Chen, X.; Schwarzschild, A.; Song, D.; Madry, A.; Li, B.; and Goldstein, T. 2023.
\newblock Dataset Security for Machine Learning: Data Poisoning, Backdoor Attacks, and Defenses.
\newblock \emph{IEEE Transactions on Pattern Analysis and Machine Intelligence}, 45(02): 1563--1580.

\bibitem[{G{{\"o}}nen and Alpaydin(2011)}]{mkl}
G{{\"o}}nen, M.; and Alpaydin, E. 2011.
\newblock Multiple Kernel Learning Algorithms.
\newblock \emph{Journal of Machine Learning Research}, 12(64): 2211--2268.

\bibitem[{Gonzalez and Woods(2006)}]{dip-book}
Gonzalez, R.~C.; and Woods, R.~E. 2006.
\newblock \emph{Digital Image Processing (3rd Edition)}.
\newblock USA: Prentice-Hall, Inc.
\newblock ISBN 013168728X.

\bibitem[{He, Zha, and Katabi(2023)}]{he2023indiscriminate}
He, H.; Zha, K.; and Katabi, D. 2023.
\newblock Indiscriminate Poisoning Attacks on Unsupervised Contrastive Learning.
\newblock In \emph{The Eleventh International Conference on Learning Representations}.

\bibitem[{He et~al.(2022)He, Chen, Xie, Li, Doll\'ar, and Girshick}]{MAE}
He, K.; Chen, X.; Xie, S.; Li, Y.; Doll\'ar, P.; and Girshick, R. 2022.
\newblock Masked Autoencoders Are Scalable Vision Learners.
\newblock In \emph{Proceedings of the IEEE/CVF Conference on Computer Vision and Pattern Recognition (CVPR)}, 16000--16009.

\bibitem[{He et~al.(2016)He, Zhang, Ren, and Sun}]{resnet}
He, K.; Zhang, X.; Ren, S.; and Sun, J. 2016.
\newblock Deep Residual Learning for Image Recognition.
\newblock In \emph{2016 CVPR}, 770--778.

\bibitem[{Hedstr{\"o}m et~al.(2024)Hedstr{\"o}m, Weber, Lapuschkin, and H{\"o}hne}]{hedstroem2023sanity}
Hedstr{\"o}m, A.; Weber, L.; Lapuschkin, S.; and H{\"o}hne, M. 2024.
\newblock A Fresh Look at Sanity Checks for Saliency Maps.
\newblock In \emph{Explainable Artificial Intelligence}, 403--420. Cham: Springer Nature Switzerland.

\bibitem[{Hein, Andriushchenko, and Bitterwolf(2019)}]{hein}
Hein, M.; Andriushchenko, M.; and Bitterwolf, J. 2019.
\newblock Why ReLU Networks Yield High-Confidence Predictions Far Away From the Training Data and How to Mitigate the Problem.
\newblock In \emph{2019 IEEE/CVF Conference on Computer Vision and Pattern Recognition (CVPR)}, 41--50. Los Alamitos, CA, USA: IEEE Computer Society.

\bibitem[{Helber et~al.(2018)Helber, Bischke, Dengel, and Borth}]{helber2018introducing}
Helber, P.; Bischke, B.; Dengel, A.; and Borth, D. 2018.
\newblock Introducing EuroSAT: A Novel Dataset and Deep Learning Benchmark for Land Use and Land Cover Classification.
\newblock In \emph{IGARSS 2018-2018 IEEE International Geoscience and Remote Sensing Symposium}, 204--207. IEEE.

\bibitem[{Kahl et~al.(2024)Kahl, L{\"u}th, Zenk, Maier-Hein, and Jaeger}]{kahl2024values}
Kahl, K.-C.; L{\"u}th, C.~T.; Zenk, M.; Maier-Hein, K.; and Jaeger, P.~F. 2024.
\newblock Val{UES}: A Framework for Systematic Validation of Uncertainty Estimation in Semantic Segmentation.
\newblock In \emph{The Twelfth International Conference on Learning Representations}.

\bibitem[{Kompa, Snoek, and Beam(2021)}]{Kompa2021}
Kompa, B.; Snoek, J.; and Beam, A.~L. 2021.
\newblock Second opinion needed: communicating uncertainty in medical machine learning.
\newblock \emph{npj Digital Medicine}.

\bibitem[{Lakshminarayanan, Pritzel, and Blundell(2017)}]{simpleensemble}
Lakshminarayanan, B.; Pritzel, A.; and Blundell, C. 2017.
\newblock Simple and scalable predictive uncertainty estimation using deep ensembles.
\newblock In \emph{Proceedings of the 31st International Conference on Neural Information Processing Systems}, NIPS'17, 6405–6416. Red Hook, NY, USA: Curran Associates Inc.
\newblock ISBN 9781510860964.

\bibitem[{Lei et~al.(2023)Lei, Li, Li, Zhang, and Shan}]{NEURIPS2023c2eac51b}
Lei, Y.; Li, Z.; Li, Y.; Zhang, J.; and Shan, H. 2023.
\newblock LICO: Explainable Models with Language-Image COnsistency.
\newblock In Oh, A.; Naumann, T.; Globerson, A.; Saenko, K.; Hardt, M.; and Levine, S., eds., \emph{Advances in Neural Information Processing Systems}, volume~36, 61870--61887. Curran Associates, Inc.

\bibitem[{Li and Lee(1993)}]{Li1993}
Li, C.; and Lee, C. 1993.
\newblock Minimum cross entropy thresholding.
\newblock \emph{Pattern Recognition}, 26(4): 617–625.

\bibitem[{Li and Tam(1998)}]{Li1998}
Li, C.; and Tam, P. 1998.
\newblock An iterative algorithm for minimum cross entropy thresholding.
\newblock \emph{Pattern Recognition Letters}, 19(8): 771–776.

\bibitem[{Lin et~al.(2023)Lin, Chen, Kim, and Lee}]{lin2023contrastive}
Lin, C.; Chen, H.; Kim, C.; and Lee, S.-I. 2023.
\newblock Contrastive Corpus Attribution for Explaining Representations.
\newblock In \emph{The Eleventh International Conference on Learning Representations}.

\bibitem[{Lin et~al.(2014)Lin, Maire, Belongie, Hays, Perona, Ramanan, Doll{\'{a}}r, and Zitnick}]{Lin2014}
Lin, T.-Y.; Maire, M.; Belongie, S.; Hays, J.; Perona, P.; Ramanan, D.; Doll{\'{a}}r, P.; and Zitnick, C.~L. 2014.
\newblock Microsoft {COCO}: Common Objects in Context.
\newblock In \emph{Computer Vision {\textendash} {ECCV} 2014}, 740--755. Springer International Publishing.

\bibitem[{Lundberg and Lee(2017)}]{kernelshap}
Lundberg, S.~M.; and Lee, S.-I. 2017.
\newblock A unified approach to interpreting model predictions.
\newblock In \emph{Proceedings of the 31st International Conference on Neural Information Processing Systems}, NIPS'17, 4768–4777. Red Hook, NY, USA: Curran Associates Inc.
\newblock ISBN 9781510860964.

\bibitem[{Maddox et~al.(2019)Maddox, Izmailov, Garipov, Vetrov, and Wilson}]{swag}
Maddox, W.~J.; Izmailov, P.; Garipov, T.; Vetrov, D.~P.; and Wilson, A.~G. 2019.
\newblock A Simple Baseline for Bayesian Uncertainty in Deep Learning.
\newblock In Wallach, H.; Larochelle, H.; Beygelzimer, A.; d\textquotesingle Alch\'{e}-Buc, F.; Fox, E.; and Garnett, R., eds., \emph{Advances in Neural Information Processing Systems}, volume~32. Curran Associates, Inc.

\bibitem[{M{\o}ller et~al.(2024)M{\o}ller, Igel, Wickstr{\o}m, Sporring, Jenssen, and Ibragimov}]{moller2024finding}
M{\o}ller, B.~L.; Igel, C.; Wickstr{\o}m, K.~K.; Sporring, J.; Jenssen, R.; and Ibragimov, B. 2024.
\newblock Finding {NEM}-U: Explaining unsupervised representation learning through neural network generated explanation masks.
\newblock In \emph{Forty-first International Conference on Machine Learning}.

\bibitem[{Morch et~al.(1995)}]{kaihansen95}
Morch, N.; et~al. 1995.
\newblock Visualization of neural networks using saliency maps.
\newblock In \emph{International Conference on Neural Networks}, 2085--2090.

\bibitem[{Otsu(1979)}]{otsu}
Otsu, N. 1979.
\newblock A Threshold Selection Method from Gray-Level Histograms.
\newblock \emph{IEEE Transactions on Systems, Man, and Cybernetics}, 9(1): 62--66.

\bibitem[{Paszke et~al.(2019)Paszke, Gross, Massa et~al.}]{pytorch}
Paszke, A.; Gross, S.; Massa, F.; et~al. 2019.
\newblock PyTorch: An Imperative Style, High-Performance Deep Learning Library.
\newblock In \emph{Advances in Neural Information Processing Systems}, 8024--8035.

\bibitem[{Petsiuk, Das, and Saenko(2018)}]{Petsiuk2018rise}
Petsiuk, V.; Das, A.; and Saenko, K. 2018.
\newblock RISE: Randomized Input Sampling for Explanation of Black-box Models.
\newblock In \emph{Proceedings of the British Machine Vision Conference (BMVC)}.

\bibitem[{Reynolds(2009)}]{Reynolds2009}
Reynolds, D. 2009.
\newblock \emph{Gaussian Mixture Models}, 659--663.
\newblock Boston, MA: Springer US.
\newblock ISBN 978-0-387-73003-5.

\bibitem[{Schulz, Santos-Rodriguez, and Poyiadzi(2022)}]{Schulz2022}
Schulz, J.; Santos-Rodriguez, R.; and Poyiadzi, R. 2022.
\newblock Uncertainty Quantification of Surrogate Explanations: an Ordinal Consensus Approach.
\newblock \emph{Proceedings of the Northern Lights Deep Learning Workshop}, 3.

\bibitem[{Sehwag, Chiang, and Mittal(2021)}]{sehwag2021ssd}
Sehwag, V.; Chiang, M.; and Mittal, P. 2021.
\newblock SSD: A Unified Framework for Self-Supervised Outlier Detection.
\newblock In \emph{International Conference on Learning Representations}.

\bibitem[{Slack et~al.(2021)Slack, Hilgard, Singh, and Lakkaraju}]{posthocreliable}
Slack, D.; Hilgard, A.; Singh, S.; and Lakkaraju, H. 2021.
\newblock Reliable Post hoc Explanations: Modeling Uncertainty in Explainability.
\newblock In Ranzato, M.; Beygelzimer, A.; Dauphin, Y.; Liang, P.; and Vaughan, J.~W., eds., \emph{Advances in Neural Information Processing Systems}, volume~34, 9391--9404. Curran Associates, Inc.

\bibitem[{Springenberg et~al.(2015)Springenberg, Dosovitskiy, Brox, and Riedmiller}]{guidebackprop}
Springenberg, J.~T.; Dosovitskiy, A.; Brox, T.; and Riedmiller, M. 2015.
\newblock Striving for Simplicity: The All Convolutional Net.
\newblock In \emph{ICLR Workshop}.

\bibitem[{Srivastava et~al.(2014)Srivastava, Hinton, Krizhevsky, Sutskever, and Salakhutdinov}]{dropout}
Srivastava, N.; Hinton, G.; Krizhevsky, A.; Sutskever, I.; and Salakhutdinov, R. 2014.
\newblock Dropout: A Simple Way to Prevent Neural Networks from Overfitting.
\newblock \emph{Journal of Machine Learning Research}, 15(56): 1929--1958.

\bibitem[{Sundararajan, Taly, and Yan(2017)}]{integratedgradients}
Sundararajan, M.; Taly, A.; and Yan, Q. 2017.
\newblock Axiomatic Attribution for Deep Networks.
\newblock In Precup, D.; and Teh, Y.~W., eds., \emph{Proceedings of the 34th International Conference on Machine Learning, {ICML} 2017, Sydney, NSW, Australia, 6-11 August 2017}, volume~70 of \emph{Proceedings of Machine Learning Research}, 3319--3328. {PMLR}.

\bibitem[{Tonekaboni et~al.(2019)Tonekaboni, Joshi, McCradden, and Goldenberg}]{tonekamboni}
Tonekaboni, S.; Joshi, S.; McCradden, M.~D.; and Goldenberg, A. 2019.
\newblock What Clinicians Want: Contextualizing Explainable Machine Learning for Clinical End Use.
\newblock In \emph{Proceedings of the 4th Machine Learning for Healthcare Conference}, volume 106, 359--380.

\bibitem[{Trosten et~al.(2023)Trosten, L{\o}kse, Jenssen, and Kampffmeyer}]{Trosten2023CVPR}
Trosten, D.~J.; L{\o}kse, S.; Jenssen, R.; and Kampffmeyer, M.~C. 2023.
\newblock On the Effects of Self-Supervision and Contrastive Alignment in Deep Multi-View Clustering.
\newblock In \emph{Proceedings of the IEEE/CVF Conference on Computer Vision and Pattern Recognition (CVPR)}, 23976--23985.

\bibitem[{Wang, Zhang, and Lim(2021)}]{Wang2021uc}
Wang, D.; Zhang, W.; and Lim, B.~Y. 2021.
\newblock Show or suppress? Managing input uncertainty in machine learning model explanations.
\newblock \emph{Artificial Intelligence}, 294: 103456.

\bibitem[{Wang et~al.(2019)Wang, Li, Aertsen, Deprest, Ourselin, and Vercauteren}]{ttadropout}
Wang, G.; Li, W.; Aertsen, M.; Deprest, J.; Ourselin, S.; and Vercauteren, T. 2019.
\newblock Aleatoric uncertainty estimation with test-time augmentation for medical image segmentation with convolutional neural networks.
\newblock \emph{Neurocomputing}, 338: 34--45.

\bibitem[{Weinberger, Lin, and Lee(2023)}]{Weinberger2023}
Weinberger, E.; Lin, C.; and Lee, S.-I. 2023.
\newblock Isolating salient variations of interest in single-cell data with contrastiveVI.
\newblock \emph{Nature Methods}, 20(9): 1336–1345.

\bibitem[{Wickstr{\o}m, Kampffmeyer, and Jenssen(2020)}]{Wickstrm2020}
Wickstr{\o}m, K.; Kampffmeyer, M.; and Jenssen, R. 2020.
\newblock Uncertainty and interpretability in convolutional neural networks for semantic segmentation of colorectal polyps.
\newblock \emph{Medical Image Analysis}, 60: 101619.

\bibitem[{Wickstr{\o}m et~al.(2021)Wickstr{\o}m, Mikalsen, Kampffmeyer, Revhaug, and Jenssen}]{9284514}
Wickstr{\o}m, K.; Mikalsen, K.~{\O}.; Kampffmeyer, M.; Revhaug, A.; and Jenssen, R. 2021.
\newblock Uncertainty-Aware Deep Ensembles for Reliable and Explainable Predictions of Clinical Time Series.
\newblock \emph{IEEE Journal of Biomedical and Health Informatics}, 25(7): 2435--2444.

\bibitem[{Wickstr{\o}m et~al.(2023)Wickstr{\o}m, Trosten, L{\o}kse, Boubekki, Mikalsen, Kampffmeyer, and Jenssen}]{relax}
Wickstr{\o}m, K.~K.; Trosten, D.~J.; L{\o}kse, S.; Boubekki, A.; Mikalsen, K.~{\O}.; Kampffmeyer, M.~C.; and Jenssen, R. 2023.
\newblock {RELAX:} Representation Learning Explainability.
\newblock \emph{Int. J. Comput. Vis.}, 1584--1610.

\bibitem[{Wickstrøm et~al.(2023)Wickstrøm, Østmo, Radiya, Øyvind Mikalsen, Kampffmeyer, and Jenssen}]{cmig-wickstrom}
Wickstrøm, K.~K.; Østmo, E.~A.; Radiya, K.; Øyvind Mikalsen, K.; Kampffmeyer, M.~C.; and Jenssen, R. 2023.
\newblock A clinically motivated self-supervised approach for content-based image retrieval of CT liver images.
\newblock \emph{Computerized Medical Imaging and Graphics}, 102239.

\bibitem[{Xiao, Rasul, and Vollgraf(2017)}]{fashionmnist}
Xiao, H.; Rasul, K.; and Vollgraf, R. 2017.
\newblock Fashion-MNIST: a Novel Image Dataset for Benchmarking Machine Learning Algorithms.
\newblock \emph{CoRR}, abs/1708.07747.

\bibitem[{Zack, Rogers, and Latt(1977)}]{Zack1977}
Zack, G.~W.; Rogers, W.~E.; and Latt, S.~A. 1977.
\newblock Automatic measurement of sister chromatid exchange frequency.
\newblock \emph{Journal of Histochemistry and amp; Cytochemistry}, 25(7): 741–753.

\bibitem[{Zhang et~al.(2019)Zhang, Song, Sun, Tan, and Udell}]{LimeUc}
Zhang, Y.; Song, K.; Sun, Y.; Tan, S.; and Udell, M. 2019.
\newblock {"Why Should You Trust My Explanation?" Understanding Uncertainty in LIME Explanations}.
\newblock In \emph{Workshop on AI for Social Good}.

\end{thebibliography}



\section*{Supplementary material}

# Supplementary Material
# REPEAT: Improving Uncertainty Estimation in Representation Learning Explainability

## Appendix A

This appendix provides an in-depth analysis and discussion of important parts of REPEAT. We review the background theory on RELAX to prepare the reader for out theoretical results. We describe the different thresholding methods examined for use in REPEAT. We provide a theoretical analysis of REPEAT that shows a connection with the kenrle learning literature. Finally, we derive a new bound that reduces the number of masks required to compute importance with RELAX.

## Background Theory on RELAX

We present the background theory for how to obtain importance and uncertainty estimates in the RELAX framework (Wickstrøm et al. 2023). Let $\mathbf{X} \in \mathbb{R}^{H \times W}$ represent an image with $H \times W$ pixels, and $f$ denote a feature extractor that transforms $\mathbf{X}$ into a representation $\mathbf{h} = f(\mathbf{X}) \in \mathbb{R}^D$. RELAX works by masking out regions of the input and measuring the similarity between the unmasked and masked representation of the input. Regions are masked out with a stochastic mask $\mathbf{M} \in [0, 1]^{H \times W}$, where each element $M_{ij}$ is drawn from some distribution. The masked representation is denoted $\bar{\mathbf{h}} = f(\mathbf{X} \odot \mathbf{M})$, where $\odot$ denotes element-wise multiplication, and the similarity between $\mathbf{h}$ and $\bar{\mathbf{h}}$ as $s(\mathbf{h}, \bar{\mathbf{h}})$. A set of $N$ masks $S_k = \{\mathbf{M}^k(1), \cdots, \mathbf{M}^k(N)\}$ is randomly generated and used to generate $N$ masked representations. The importance of a pixel $R_{ij}^{\text{rlx}}$ can be computed as:

$$\bar{R}_{ij}^{\text{rlx}}(k) = \frac{1}{N} \sum_{n=1}^{N} s(\mathbf{h}, \bar{\mathbf{h}}_n) M_{ij}^k(n), \quad (1)$$

where $\bar{\mathbf{h}}_n$ is the representation of the image masked with mask $n$ and $M_{ij}(n)$ the value of element $(i, j)$ for mask $n$ and belonging to set $k$ of random masks. The cosine similarity is used as the similarity measure. An important contribution of RELAX is that it also estimates the uncertainty in the importance estimate:

$$\bar{U}_{ij}^{\text{rlx}}(k') = \frac{1}{N} \sum_{n=1}^{N} (s(\mathbf{h}, \bar{\mathbf{h}}_n) - R_{ij}^{\text{rlx}}(k))^2 M_{ij}^k(n). \quad (2)$$

Eq. 1 assigns an importance score to each input pixel, and Eq. 2 measures the variability of the importance scores from Eq. 1. In other words, Eq. 2 tells us the variation of the particular value of the importance score. *But it does not tell us if a pixel is certainly important.* This distinction is important, and we hypothesis that a framework based on determining if a pixel is important or not with certainty, as opposed to assigning a score to a pixel, will lead to better uncertainty estimates that align more with human intuition.

## Setting the Threshold

Finding a proper value for $\tau$ in Equation 3 is crucial to generate meaningful samples. We propose to approach this from a foreground-background thresholding perspective, where we consider important pixels as foreground pixels and unimportant pixels as background pixels. Thresholding is a classic problem in image processing with a vast amount of established literature (Gonzalez and Woods 2006). Thresholding algorithms can be separated into two categories; histogram-based or local methods. Histogram-based methods use the histogram of pixel intensities, while local methods process each pixel by considering its neighborhood. Local methods are more computationally demanding, and since we will repeat the thresholding procdure multiple times, we will only focus on histogram-based methods. We consider the following four thresholding methods as potential candidates to use in REPEAT:

- **Mean**: Using the mean of the histogram is a simple technique to set the threshold between foreground and background, but can often be powerful and provide competitive performance with more sophisticated algorithms (Glasbey 1993).
- **Otsu**: Otsu's method minimizes the intra-class variance between two classes (foreground and background) (Otsu 1979), and is one of the fundamental techniques within automatic image thresholding. Otus's method typically performs well if the histogram has a bimodal distribution.
- **Triangle**: The triangle algorithm creates a triangle using the maximum peak and final histogram bin, and leverages this triangle to fin the histogram bin the the greatest distance between the top and the hypotenuse of the triangle (Zack, Rogers, and Latt 1977). This method is designed with unimodal histograms in mind.
- **Li**: Li's method leverages gradient descent to iteratively minimize a cross entropy loss (Li and Lee 1993; Li and

Tam 1998). Due to the gradients descent-based optimization, Li's method can be vulnerable to local minima if the histogram has many modes.

## Connecting REPEAT with the Kernel Learning literature

REPEAT tells us the probability of a pixel being important for the representation of an image and of how certain we are of importance. However, here we present a novel analysis which shows that REPEAT can be understood from the perspective of ensembles (Hastie, Tibshirani, and Friedman 2008) and multiple kernel learning (Gönen and Alpaydin 2011). This is an important contribution, as it opens up possibilities to connect REPEAT with a greater base of existing literature, which can be used in future developments of REPEAT. First, we assume that the base stochastic R-XAI method used in REPEAT is the RELAX framework (Wickstrøm et al. 2023). RELAX is a masking-based approach that removes pixels using a mask ($\mathbf{M}$) and measures the similarity between the unmasked ($\mathbf{h}$) and masked representation $\bar{\mathbf{h}}$. This procedure is performed for $N$ masks, resulting in a estimate for the importance of a pixel for the representation of an image. With certain assumptions on the similarity measure $s(\cdot, \cdot)$, the importance score can be understood as a weighted linear scoring function in a reproducing kernel Hilbert space (RKHS) in a multiple kernel learning setting (Gönen and Alpaydin 2011):

**Theorem 1** Suppose the similarity function $s(\cdot, \cdot)$ is a valid Mercer kernel (Mercer 1909). The importance $\bar{R}_{ij}$ acts as a conic sum of weighted linear scoring functions between $\mathbf{h}$, and the weighted mean of $\bar{\mathbf{h}}_1, \ldots, \bar{\mathbf{h}}_N$ in the RKHS induced by $s(\cdot, \cdot)$. That is:

$$\bar{R}_{ij} = \frac{1}{K}\sum_{k=1}^{K} \gamma_k \langle \phi(\mathbf{h}), \frac{1}{N}\sum_{n=1}^{N} \phi(\bar{\mathbf{h}}_n) M_{ij}^k(n) \rangle_{\mathcal{H}} \quad (3)$$

where $\phi : \mathbb{R}^d \to \mathcal{H}$ is the mapping to the RKHS, $\mathcal{H}$, induced by the kernel $s(\cdot, \cdot)$, $\gamma_k = I_{ij}(k)/A(k)$, and $\langle \cdot, \cdot \rangle_{\mathcal{H}}$ is the inner product on $\mathcal{H}$.

Here, $M_{ij}^k(n)$ is the masking value for pixel $(i, j)$ in the $k^{\text{th}}$ repeat for mask $n$. See proof below. Th. 1 provides a deeper understanding of the underlying mechanism of REPEAT, since the importance score can be understood from the perspective of kernel learning (Kung 2014) and linear models (McCullagh and Nelder 2019). By connecting REPEAT with existing literature it allows us to potentially use insights from the established literature to improve REPEAT. For instance, techniques from unsupervised multiple kernel learning (Zhuang et al. 2011) could be considered to improve the combination of realizations of the Bernoulli RVs, or methods for regularizing the kernels could be incorporated (Varma and Babu 2009). We consider such directions as exciting research directions for the future, but outside the scope of this work.

**Proof of Theorem 1**
Since $s(\cdot, \cdot)$ is a valid Mercer kernel, we can write $s(\mathbf{h}, \bar{\mathbf{h}}_n) = \langle \phi(\mathbf{h}), \phi(\bar{\mathbf{h}}_n) \rangle_{\mathcal{H}}$. This gives

$$\bar{R}_{ij}^{\text{rlx}}(k) = \frac{1}{N}\sum_{n=1}^{N} \langle \phi(\mathbf{h}), \phi(\bar{\mathbf{h}}_n) \rangle_{\mathcal{H}} M_{ij}^k(n) \quad (4)$$

$$= \langle \phi(\mathbf{h}), \frac{1}{N}\sum_{n=1}^{N} \phi(\bar{\mathbf{h}}_n) M_{ij}^k(n) \rangle_{\mathcal{H}} \quad (5)$$

by the bilinearity of the inner product on $\mathcal{H}$.

Then, the functional form of a multiple kernel learning algorithm is defined as:

**Definition of Functional form of Multiple Kernel Learning Algorithm** The functional form of a multiple kernel learning algorithm with a weighted linear combination function can be expressed as:

$$\kappa_{\gamma_k}(x, x') = \sum_{k=1}^{K} \gamma_k \kappa_k(x, x') \quad (6)$$

Then, remember that the explanation of REPEAT are defined as:

$$\bar{R}_{ij} = \frac{1}{K}\sum_{k=1}^{K} I_{ij}(k) W_{ij}(k), \quad (7)$$

where

$$W_{ij}(k) = \frac{\bar{R}_{ij}^{\text{base}}(k)}{A(k)} \quad (8)$$

and $A(k)$ is the maximum value of $\bar{\mathbf{R}}^{\text{base}}(k)$. Assuming that $\bar{\mathbf{R}}^{\text{base}}(k) = \bar{\mathbf{R}}^{\text{rlx}}(k)$, and that $I_{ij}/A(k) = \gamma_k$, Eq. 7 can be expressed as:

$$\bar{R}_{ij} = \frac{1}{K}\sum_{k=1}^{K} \gamma_k \bar{\mathbf{R}}^{\text{rlx}}(k). \quad (9)$$

Combining Eq. 9 with Eq. 5, it is straight forward to obtain Eq. 3, which concludes our proof.

## New Bound for Reducing the Number of Required Masks

RELAX uses 3000 masks to generate its importance scores. This procedure must be repeated $K$ times, and can result in REPEAT being computationally demanding. Here, we present two arguments for reducing the number of masks, thus reducing the computational demand of REPEAT.

**Argument I: New empirical bound on estimation error**
The number of masks required to create the RELAX importance scores is determined by a bound on the estimation error (Wickstrøm et al. 2023). This bound assumes that the similarity scores between masked and unmasked representations is bounded between 0 and 1. While this is true in theory, in practice the similarity scores will very rarely reach a value of 0 or 1. This is because the masking procedure will

remove large parts of the image (which reduces the possibility similarity=1) but not all parts of the image (keeping some information which reduces the possibility of similarity=0). Therefore, we propose a new bound where we allow a more flexible maximum and minimum similarity assumption:

**Theorem 2** Suppose $s(\cdot, \cdot)$ is bounded in $(0, 1)$, and $b_m$ and $a_m$ is the maximum and minimum value of $s$, respectively. Then, for any $\delta \in (0,1)$ and $t > 0$, if $N$ in Eq. 1, satisfies:

$$N \geq -(b_m - a_m)\frac{\ln(\delta/2)}{2t^2}, \quad (10)$$

we have $P(|\bar{R}_{ij} - R_{ij}| \geq t) \leq \delta$.

Th. 2 estimates the number of masks required to obtain an estimaton error lower than $t$ with a probability of $1 - \delta$, which is weighted by the difference between the maximum and minimum similarity score. If the difference between the maximum and minimum similarity score is small, the number of masks required to obtained the desired estimator error is reduced.

**Proof of Theorem 2** First, let the Bounded difference assumption be defined as follows:

**Definition of Bounded difference assumption** Let $A$ be some set and $f : A^N \to \mathbb{R}$. The function $f$ satisfies the bounded differences assumption if there exists real numbers $c_1, \ldots, c_N \geq 0$ so that for all $i = 1, \ldots, N$,

$$\sup_{x_1,\ldots,x_N, x_i \in A} |f(x_1, \ldots, x_N, x'_i) - f(x_1, \ldots, x_N, x'_i)| \quad (11)$$

We then have the following lemmas:

**Lemma 1 (McDiarmid's inequality)** Let $X_1, \ldots, X_N$ be arbitrary independent random variables on set A and $f : A^N \to \mathbb{R}$ satisfies the bounded difference assumption. Then, for all $t > 0$

$$P(|f(X_1, \ldots, X_N) - \mathbb{E}[f(X_1, \ldots, X_N)]| \geq t)$$
$$\leq 2e^{\frac{-2t^2}{\sum_{n=1}^{N} c_n^2}} \quad (12)$$

See (McDiarmid 1989) for proof.

**Lemma 2** Let $X_1, \ldots, X_N$ and $f$ be defined as in Lem. 1, then if each $X_n$ satisfies $X_n \in (a_n, b_n)$ and $f(X_1, \ldots, X_N) = \sum_{n=1}^{N} X_n$, then $c_n = b_n - a_n$.

See (McDiarmid 1989) for proof.

We are now ready to prove the theorem. First, let

$$X_n = \frac{s(\mathbf{h}, \bar{\mathbf{h}}_n)M_{ij}(n)}{N}, \quad (13)$$

and

$$f(X_1, \ldots, X_n) = \sum_{n=1}^{N} X_n. \quad (14)$$

Next, we assume that $a_m$ is the maximum value across all $a_n$ and that $b_m$ is the maximum value across all $b_n$. Since $s(\cdot, \cdot)$ is bounded in $(0, 1)$ (we use the cosine similarity between vectors with non-negative elements (ReLU outputs)), $a_m$ and $b_m$ will also be bounded $(0, 1)$.

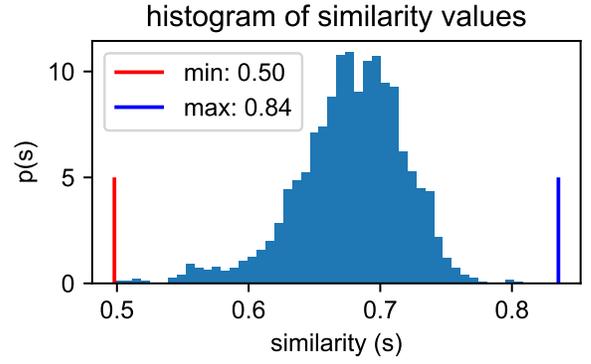

Figure 1: Similarity scores between masked and unmasked representation of 1000 randomly sampled images from Pascal VOC (Everingham et al. 2009) and embedded using a ResNet18 feature extractor (He et al. 2016).

Now, observe that

$$f(X_1, \ldots, X_n) = \frac{1}{N}\sum_{n=1}^{N} s(\mathbf{h}, \bar{\mathbf{h}}_n)M_{ij}(n) = \bar{R}_{ij}. \quad (15)$$

Combining Lem. 1 and 2 then gives

$$P(|\bar{R}_{ij} - R_{ij}|] \geq t) \leq 2e^{\frac{-2t^2}{\sum_{n=1}^{N}(\frac{b_m - a_m}{N})^2}} \quad (16)$$

for all $t > 0$. Inserting $N = -(b_m - a_m)\ln(\delta/2)/2t^2$ gives

$$P(|\bar{R}_{ij} - R_{ij}|] \geq t) \leq 2e^{\frac{-2t^2}{\sum_{n=1}^{N}(\frac{b_m - a_m}{N})^2}} \quad (17)$$
$$= 2e^{-2t^2\left(-\frac{\ln(\delta/2)}{2t^2}\right)} \quad (18)$$
$$= 2e^{\ln(\delta/2)} \quad (19)$$
$$= \delta, \quad (20)$$

which concludes our proof.

**Argument II: Diversity in importance scores** REPEAT can be thought of as an ensemble approach, where an ensemble of scoring functions are combined to provide an estimate of importance (see Th. 1. A key ingredient in ensemble learning is diversity in the ensemble members (Hastie, Tibshirani, and Friedman 2008; Mao et al. 2011). One common way in ensemble learning to increase diversity is to add noise into the development of each ensemble member (Hastie, Tibshirani, and Friedman 2008; Nordhaug Myhre et al. 2018). This can be accomplished also in our case by reducing the number of masks, thus allowing more error in each run of RELAX which would increase the diversity in the importance scores across the K repeats.

**Reducing the number of masks** By the arguments above we can reduce the number of masks. First, we empirically evaluate the maximum and minimum similarity scores obtained on 1000 randomly sampled images from Pascal VOC (Everingham et al. 2009). For each image, we generate 1000 randomly masked versions and compute the similarity between the masked and unmasked representation from a

|  | REPEAT | | |
|---|---|---|---|
| threshold method | L (↑) | C (↓) | R (↓) |
| otsu | **0.708** | 10.101 | 0.363 |
| li | 0.704 | 9.986 | <u>0.213</u> |
| triangle | 0.704 | **9.691** | 0.216 |
| mean | <u>0.706</u> | 10.076 | **0.211** |

Table 1: Results for REPEAT for localization (L), complexity (C), and robustness (R) metrics on PASCAL-VOC with an AlexNet encoder (Krizhevsky, Sutskever, and Hinton 2012) and different threshold methods in REPEAT. Best and second best performance for each column is indicated by **bold** and <u>underlined</u>, respectively.

ResNet18 feature extractor (He et al. 2016). These results are shown in Fig. 1 and the maximum and minimum similarity score (corresponding to $b_m$ and $a_m$ in Th. 2, respectively) were $b_m = 0.84$ and $a_m = 0.5$. If we want a similar estimator error as in previous work (Wickstrøm et al. 2023) ($t > 0.035$ with a probability of 0.95), then according to Th. 2 we need 1258 masks. Then, by argument number II, we allow a bigger estimation error ($t = 0.04$) to increase diversity, which reduced the number of masks further down to $\sim 1000$.

## Appendix B

This appendix provides supporting results for the main paper. We evaluate different thresholding method to investigate their suitability with REPEAT. We investigate the number of repeats necessary for good performance and how the results of REPEAT different between different runs. Further, we conduct OOD detection with Fashion MNIST as the OOD dataset, and provide quantitative and qualitative results for the explanations from REPEAT. Lastly, we evaluate REPEAT on a downstream task where uncertainty is used to detect poisoned data in the unsupervised representation learning setting (He, Zha, and Katabi 2023).

### Evaluating the Threshold Methods

To examine the threshold method used in REPEAT we consider three well-known metrics from the XAI literature, localization (Theiner, Müller-Budack, and Ewerth 2021), complexity (Bhatt, Weller, and Moura 2021), and robustness (Alvarez-Melis and Jaakkola 2018). Tab. 1 shows the results of this analysis on the importance scores of REPEAT, and we find that mean thresholding gives slightly better performance than the other methods.

### Number of Repeats (K)

To examine the number of repeats (K) used in REPEAT we consider three well-known metrics from the XAI literature, localization (Theiner, Müller-Budack, and Ewerth 2021), complexity (Bhatt, Weller, and Moura 2021), and robustness (Alvarez-Melis and Jaakkola 2018). Tab. 2 shows the results of this analysis on the importance scores of REPEAT, and we find that REPEAT is relatively stable for different numbers of K.

|  | REPEAT | | |
|---|---|---|---|
| number of repeats | L (↑) | C (↓) | R (↓) |
| 10 | **0.70** | **10.10** | <u>0.21</u> |
| 20 | **0.70** | **10.10** | 0.14 |
| 30 | **0.70** | **10.10** | **0.12** |

Table 2: Results for REPEAT for localization (L), complexity (C), and robustness (R) metrics on PASCAL-VOC with an AlexNet encoder (Krizhevsky, Sutskever, and Hinton 2012) and different number of repeats. Best and second best performance for each column is indicated by **bold** and <u>underlined</u>, respectively.

### Standard Deviation for Repeated runs of REPEAT and RELAX

We compute the complexity scores for RELAX and REPEAT based on 3 repeated runs to investigate the effect of the stochasticity in these methods. Tab. 3 shows these results, and illustrates that both RELAX and REPEAT are stable with regards to the stochasticity inherent in these methods.

### OOD detection with FashionMNIST

Tab. 4 shows the results of the OOD experiment with FashionMNIST as the OOD data. Both REPEAT and Integrated Gradients perform with a ResNet50 encoder, but all methods struggle with a ViT encoder. We hypothesis the poor performance is related to the upsampling procedure that is necessary to process FashionMNIST with a these encoders. Each image in FashioMNIST is originally 28x28 pixels, but is upsampled to a size of 224x224. Many pixels are just black background, which for the patching of the ViT will be indicated as being confidently unimportant and thus result in a low uncertainty despite it being OOD.

### Quantitative and Qualitative Results for REPEAT Explanations

Here, we present more qualitative results for REPEAT and competing methods, which are shown in Figure 2, 3, and 4. We also present quantitative results for the explanations produced by REPEAT in Table 5 and 6. The quantitative results follow similar setups as prior works (Wickstrøm et al. 2023), and we measure localization (Theiner, Müller-Budack, and Ewerth 2021), complexity (Chalasani et al. 2020), and robustness (Alvarez-Melis and Jaakkola 2018).

### Poisoned Data Detection with Uncertainty Estimates

Motivated by REPEAT's high performance in OOD detection and the close connection between reliability and XAI, we hypothesis that REPEAT can be used to identify poisoned data by treating poison detection (Goldblum et al. 2023) as an OOD problem, similar to the protocol outlined in the main paper. Existing defenses against poisoning typically rely on modifying the training procedure (Goldblum et al. 2023), but here we take a new direction and intercept

| explainability method | resnet50 | vit |
|---|---|---|
| RELAX | 10.82 ± 8.9e-06 | 10.82 ± 9.0e-06 |
| REPEAT | 9.97 ± 4.7e-04 | 9.99 ± 1.3e-03 |

| explainability method | resnet50 | vit |
|---|---|---|
| RELAX | 10.75 ± 7.2e-06 | 10.74 ± 7.5e-06 |
| REPEAT | 9.91 ± 4.4e-04 | 9.94 ± 9.8e-04 |

Table 3: Results for R-XAI method for complexity (lower is better) of uncertainty estimates on PASCAL-VOC (top) and MS-COCO (bottom) with a ResNet50 and a ViT encoder.

| | resnet50 | | vit | |
|---|---|---|---|---|
| explainability method | pascal-voc | ms-coco | pascal-voc | ms-coco |
| saliency* | 0.000 | 0.000 | 0.001 | 0.000 |
| integrated gradients* | 0.996 | **1.000** | **0.016** | **0.019** |
| guided backpropagation* | 0.146 | 0.068 | 0.002 | 0.001 |
| RELAX | 0.000 | 0.000 | 0.000 | 0.009 |
| REPEAT | **0.999** | **1.000** | 0.001 | 0.001 |

Table 4: Results for R-XAI methods for OOD detection using FashionMNIST as the OOD dataset. The table shows AUROC when classifying in-domain (VOC or COCO) vs out-of-domain clusters using a Gaussian mixture model. Best and second best performance for each column is indicated by **bold** and underlined. The * highlights that these methods are adapted to the R-XAI setting using Label-Free Feature Importance (Crabbé and van der Schaar 2022).

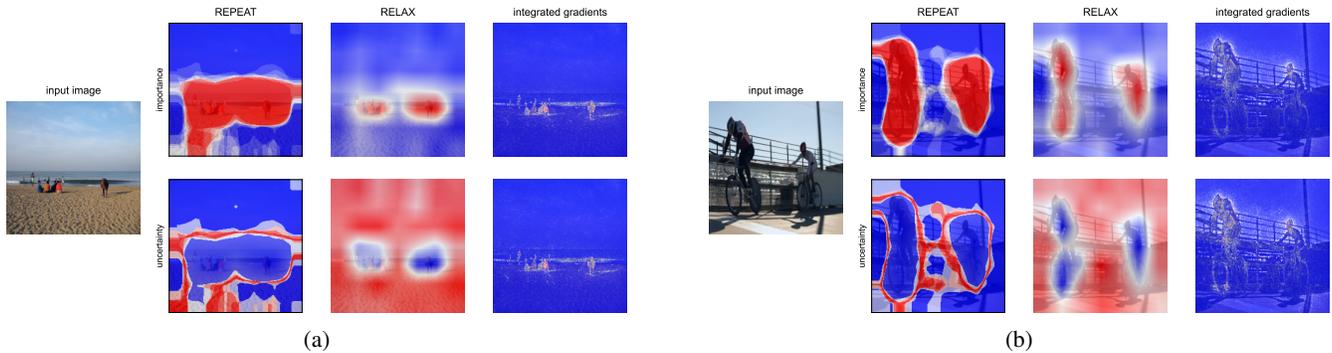

Figure 2: Qualitative examples on images from PASCAL VOC.

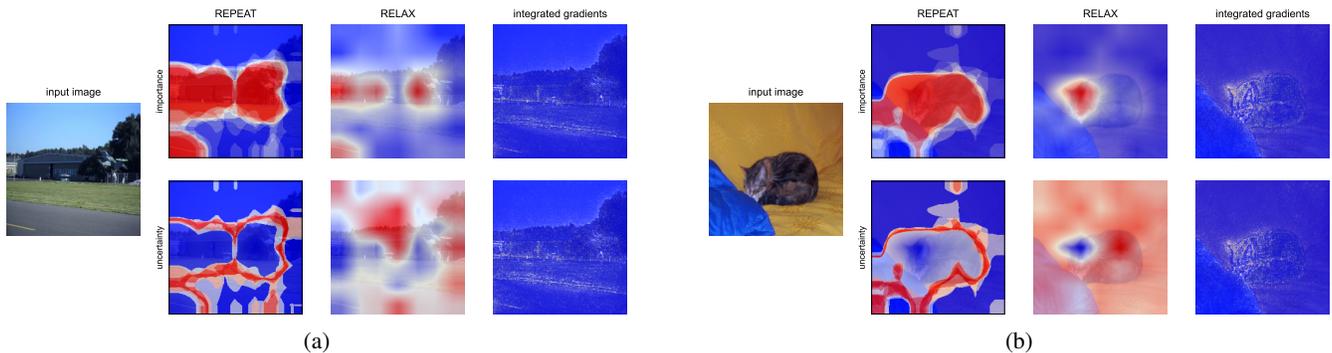

Figure 3: Qualitative examples on images from PASCAL VOC.

the poisoned data before training occurs. Specifically, we consider the case where parts of the dataset is poisoned (He, Zha, and Katabi 2023) and use the precomputed poisons for CIFAR-10 (Krizhevsky 2009) provided by He *et al.* (He,

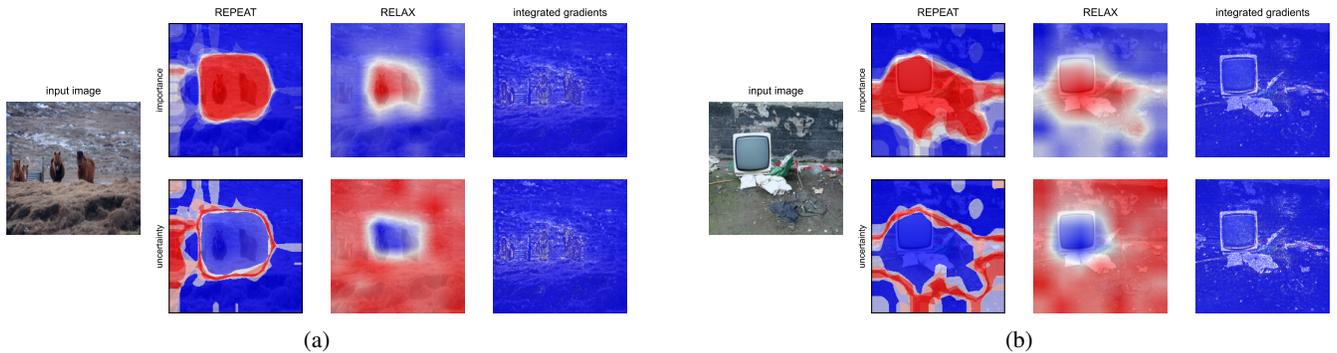

Figure 4: Qualitative examples on images from PASCAL VOC.

| | resnet50 | | | vit | | |
|---|---|---|---|---|---|---|
| explainability method | L (↑) | C (↓) | R (↓) | L (↑) | C (↓) | R (↓) |
| saliency* | 0.47 | 10.37 | 0.71 | 0.34 | 10.18 | 0.51 |
| kernel shap* | 0.29 | 10.10 | 1.51 | 0.29 | 10.29 | 1.54 |
| guided backpropagation* | 0.67 | **9.96** | <u>0.15</u> | 0.34 | 10.18 | 0.51 |
| integrated gradients* | 0.52 | 10.23 | 0.52 | <u>0.41</u> | 10.14 | 0.24 |
| RELAX | **0.71** | 10.75 | **0.01** | **0.71** | 10.75 | **0.01** |
| REPEAT | **0.71** | <u>10.10</u> | 0.22 | **0.71** | **10.06** | <u>0.22</u> |

Table 5: Results for representation learning XAI methods for localization (L), complexity (C), and robustness (R) metrics on MS-COCO with a ResNet50 and a ViT encoder. Best and second best performance for each column is indicated by **bold** and <u>underlined</u>, respectively. The * highlights that these XAI methods are adapted to the representation learning XAI setting using Label-Free Feature Importance (Crabbé and van der Schaar 2022).

Zha, and Katabi 2023). We embed the data using a pretrained ResNet18 and compute the aggregated importance and uncertainty scores using REPEAT. We consider the case introduced by He *et al.* (He, Zha, and Katabi 2023) where only a portion of the dataset is poisoned. For example, if we consider 90% of the data to be poisoned, we sample 900 poisoned images and 100 different non-poisoned images.

**Detecting poison with REPEAT:** Tab. 7 displays the results of detection scheme outlined above where difference percentages of the data is poisoned. When the mixture model is trained on aggregated importance score, the performance is low. In contrast, training on the aggregated uncertainty scores gives high performance across all percentages, thus providing an effective unsupervised poison detector. Fig. 5 shows an example of how the distribution of aggregated importance and uncertainty scores looks in the 50% poison setting. This example clearly shows how the aggregated importance scores for poisoned and non-poisoned data overlap, while the aggregated uncertainty scores shows a clear difference between the two distributions. Fig. 5 also shows the receiver operating characteristic (ROC) curve for the models trained on aggregated importance and uncertainty scores.

**How does uncertainty find poison:** The results in Tab. 7 show that the uncertainty estimates from REPEAT can reliably identify poisoned data in the unsupervised setting. But why does it work? We illuminate this question by presenting an example of the same image with or without poison shown in Fig. 6. For both the clean and poisoned image, the bird is highlighted as the most important part for the representation of the image. However, the uncertainty estimates are very different, and reveal the subtle pattern of the poison. We hypothesis that the masking process interferes with the poison pattern which leads to importance of poisoned pixels to fluctuate between being highly important or not important at all, which is exactly what the uncertainty estimates measure.

# Appendix C

This appendix present our answers to the reproducibility checklist.

- Includes a conceptual outline and/or pseudocode description of AI methods introduced: Yes.
- Clearly delineates statements that are opinions, hypothesis, and speculation from objective facts and results: Yes.
- Provides well marked pedagogical references for less-familiare readers to gain background necessary to replicate the paper: Yes.
- Does this paper make theoretical contributions: Yes.
- All assumptions and restrictions are stated clearly and formally: Yes.
- All novel claims are stated formally (e.g., in theorem statements): Yes.
- Proofs of all novel claims are included: Yes.
- Proof sketches or intuitions are given for complex and/or novel results: Yes.

| explainability method | resnet50 | | | vit | | |
|---|---|---|---|---|---|---|
| | L (↑) | C (↓) | R (↓) | L (↑) | C (↓) | R (↓) |
| saliency* | 0.62 | 10.44 | 0.73 | 0.47 | 10.24 | 0.54 |
| kernel shap* | 0.29 | <u>10.09</u> | 1.50 | 0.44 | 10.36 | 1.58 |
| guided backpropagation* | **0.86** | **10.02** | <u>0.15</u> | 0.47 | 10.24 | 0.54 |
| integrated gradients* | <u>0.67</u> | 10.29 | 0.52 | <u>0.55</u> | <u>10.2</u> | 0.25 |
| RELAX | **0.86** | 10.82 | **0.01** | **0.85** | 10.82 | **0.01** |
| REPEAT | **0.86** | 10.17 | 0.22 | **0.85** | **10.12** | <u>0.21</u> |

Table 6: Results for representation learning XAI methods for localization (L), complexity (C), and robustness (R) metrics on PASCAL-VOC with a ResNet50 and a ViT encoder. Best and second best performance for each column is indicated by **bold** and <u>underlined</u>, respectively. The * highlights that these XAI methods are adapted to the representation learning XAI setting using Label-Free Feature Importance (Crabbé and van der Schaar 2022).

| aggregated score | Percentage of poisoned samples | | | | | | | | |
|---|---|---|---|---|---|---|---|---|---|
| | 90% | 80% | 70% | 60% | 50% | 40% | 30% | 20% | 10% |
| importance | <u>0.56</u> | <u>0.62</u> | <u>0.62</u> | <u>0.59</u> | <u>0.59</u> | <u>0.58</u> | <u>0.58</u> | <u>0.59</u> | <u>0.60</u> |
| uncertainty | **0.85** | **0.86** | **0.87** | **0.86** | **0.86** | **0.85** | **0.84** | **0.85** | **0.85** |

Table 7: AUROC for poisoned data detection on CIFAR-10 using a Gaussian mixture model on either aggregated importance or uncertainty scores. The mixture models are trained on numerous cases where the percentage of poisoned samples in the dataset is different. Best and second best performance for each column is indicated by **bold** and <u>underlined</u>. The results show that the aggregated uncertainty scores are effective at identifying poisoned data, in contrast to aggregated importance scores.

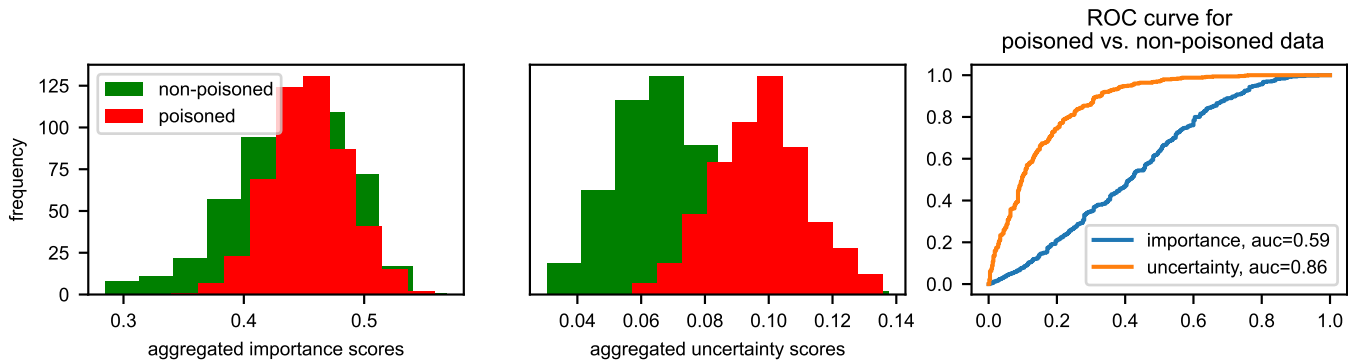

Figure 5: Histogram of aggregated importance (left) and uncertainty (middle) for poisoned and non-poisoned data from CIFAR-10. This example illustrates how the distributions are not separable based importance but clearly separable based on uncertainty, highlighting the significance of uncertainty modeling. The right plot shows the ROC curve for a Gaussian mixture model for detection of poisoned data based on importance or uncertainty scores, where the uncertainty-based model gets better performance.

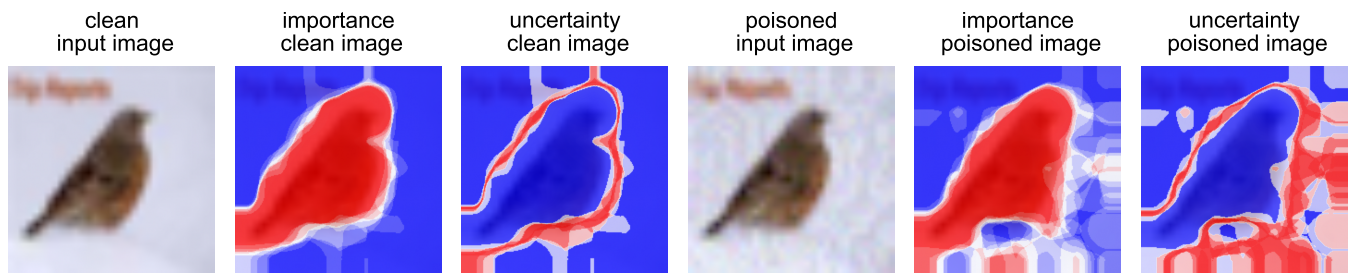

Figure 6: Importance and uncertainty for the same image from CIFAR-10 but without and with poisoning. While the importance scores are mostly similar for the non-poisoned and poisoned image, the uncertainty reveals issues in the poisoned case.

- Appropriate citations to theoretical tools used are given: Yes.
- All theoretical claims are demonstrated empirically to hold: Yes.
- All experimental code used to eliminate or disprove claims is included: Yes.
- Does this paper rely on one or more datasets? Yes.
- A moivation is given for why the experiments are conducted on the selected datasets: Yes
- All novel datasets introduced in this paper are included in a data appendix: NA.
- All novel datasets introduced in this paper will be made publicly available upon publication of the paper with a license that allows free usage for research purposes: NA.
- All datasets drawn from the existing literature (potentially including authors' own previously published work) are accompanied by appropriate citations: Yes.
- All datasets drawn from the existing literature (potentially including authors' own previously published work) are publicly available.: Yes.
- All datasets that are not publicly available are described in detail, with explanation why publicly available alternatives are not scientifically satisfising: NA.
- Does this paper include computational experiments? Yes.
- Any code required for pre-processing data is included in the appendix: Yes.
- All source code required for conducting and analyzing the experiments is included in a code appendix: Yes.
- All source code required for conducting and analyzing the experiments will be made publicly available upon publication of the paper with a license that allows free usage for research purposes: Yes.
- All source code implementing new methods have comments detailing the implementation, with references to the paper where each step comes from: Yes.
- If an algorithm depends on randomness, then the method used for setting seeds is described in a way sufficient to allow replication of results: Yes.
- This paper specifies the computing infrastructure used for running experiments (hardware and software), including GPU/CPU models; amount of memory; operating system; names and versions of relevant software libraries and frameworks: Yes.
- This paper formally describes evaluation metrics used and explains the motivation for choosing these metrics: Yes.
- This paper states the number of algorithm runs used to compute each reported result: Yes.
- Analysis of experiments goes beyond single-dimensional summaries of performance (e.g., average; median) to include measures of variation, confidence, or other distributional information: Yes.
- The significance of any improvement or decrease in performance is judged using appropriate statistical tests (e.g., Wilcoxon signed-rank): Yes.
- This paper lists all final (hyper-)parameters used for each model/algorithm in the paper's experiments: Yes.
- This paper states the number and range of values tried per (hyper-) parameter during development of the paper, along with the criterion used for selecting the final parameter setting: Yes.

# References

Alvarez-Melis, D.; and Jaakkola, T. S. 2018. Towards Robust Interpretability with Self-Explaining Neural Networks. In *Proceedings of the 32nd International Conference on Neural Information Processing Systems*, NIPS'18, 7786–7795. Red Hook, NY, USA: Curran Associates Inc.

Bhatt, U.; Weller, A.; and Moura, J. M. F. 2021. Evaluating and aggregating feature-based model explanations. In *Proceedings of the Twenty-Ninth International Joint Conference on Artificial Intelligence*, IJCAI'20. ISBN 9780999241165.

Chalasani, P.; Chen, J.; Chowdhury, A. R.; Wu, X.; and Jha, S. 2020. Concise Explanations of Neural Networks using Adversarial Training. In III, H. D.; and Singh, A., eds., *Proceedings of the 37th International Conference on Machine Learning*, volume 119 of *Proceedings of Machine Learning Research*, 1383–1391. PMLR.

Crabbé, J.; and van der Schaar, M. 2022. Label-Free Explainability for Unsupervised Models. In *Proceedings of the 39th International Conference on Machine Learning*, Proceedings of Machine Learning Research, 4391–4420. PMLR.

Everingham, M.; Gool, L. V.; Williams, C. K. I.; Winn, J.; and Zisserman, A. 2009. The Pascal Visual Object Classes (VOC) Challenge. *International Journal of Computer Vision*, 303–338.

Glasbey, C. 1993. An Analysis of Histogram-Based Thresholding Algorithms. *CVGIP: Graphical Models and Image Processing*, 55(6): 532–537.

Goldblum, M.; Tsipras, D.; Xie, C.; Chen, X.; Schwarzschild, A.; Song, D.; Madry, A.; Li, B.; and Goldstein, T. 2023. Dataset Security for Machine Learning: Data Poisoning, Backdoor Attacks, and Defenses. *IEEE Transactions on Pattern Analysis and Machine Intelligence*, 45(02): 1563–1580.

Gönen, M.; and Alpaydin, E. 2011. Multiple Kernel Learning Algorithms. *Journal of Machine Learning Research*, 12(64): 2211–2268.

Gonzalez, R. C.; and Woods, R. E. 2006. *Digital Image Processing (3rd Edition)*. USA: Prentice-Hall, Inc. ISBN 013168728X.

Hastie, T.; Tibshirani, R.; and Friedman, J. 2008. *Ensemble Learning*, 605–624. Springer New York. ISBN 9780387848587.

He, H.; Zha, K.; and Katabi, D. 2023. Indiscriminate Poisoning Attacks on Unsupervised Contrastive Learning. In *The Eleventh International Conference on Learning Representations*.

He, K.; Zhang, X.; Ren, S.; and Sun, J. 2016. Deep Residual Learning for Image Recognition. In *2016 CVPR*, 770–778.

Kingma, D. P.; and Welling, M. 2014. Auto-Encoding Variational Bayes. arXiv:1312.6114.

Krizhevsky, A.; Sutskever, I.; and Hinton, G. E. 2012. ImageNet Classification with Deep Convolutional Neural Networks. In Pereira, F.; Burges, C.; Bottou, L.; and Weinberger, K., eds., *Advances in Neural Information Processing Systems*, volume 25. Curran Associates, Inc.

Kung, S. 2014. *Kernel Methods and Machine Learning*. Cambridge University Press. ISBN 9781139049221.

Li, C.; and Lee, C. 1993. Minimum cross entropy thresholding. *Pattern Recognition*, 26(4): 617–625.

Li, C.; and Tam, P. 1998. An iterative algorithm for minimum cross entropy thresholding. *Pattern Recognition Letters*, 19(8): 771–776.

Mao, S.; Jiao, L.; Xiong, L.; and Gou, S. 2011. Greedy optimization classifiers ensemble based on diversity. *Pattern Recognition*, 44(6): 1245–1261.

McCullagh, P.; and Nelder, J. 2019. *Generalized Linear Models*. Routledge. ISBN 9780203753736.

McDiarmid, C. 1989. On the method of bounded differences, 148–188. Cambridge University Press.

Mercer, J. 1909. Functions of positive and negative type, and their connection with the theory of integral equations. *Philosophical Transactions of the Royal Society, London*, 209: 415–446.

Nordhaug Myhre, J.; Øyvind Mikalsen, K.; Løkse, S.; and Jenssen, R. 2018. Robust clustering using a kNN mode seeking ensemble. *Pattern Recognition*, 76: 491–505.

Otsu, N. 1979. A Threshold Selection Method from Gray-Level Histograms. *IEEE Transactions on Systems, Man, and Cybernetics*, 9(1): 62–66.

Theiner, J.; Müller-Budack, E.; and Ewerth, R. 2021. Interpretable Semantic Photo Geolocalization. *CoRR*, abs/2104.14995.

Varma, M.; and Babu, B. R. 2009. More generality in efficient multiple kernel learning. In *Proceedings of the 26th Annual International Conference on Machine Learning*, ICML '09, 1065–1072. New York, NY, USA: Association for Computing Machinery. ISBN 9781605585161.

Wickstrøm, K. K.; Trosten, D. J.; Løkse, S.; Boubekki, A.; Mikalsen, K. Ø.; Kampffmeyer, M. C.; and Jenssen, R. 2023. RELAX: Representation Learning Explainability. *Int. J. Comput. Vis.*, 1584–1610.

Zack, G. W.; Rogers, W. E.; and Latt, S. A. 1977. Automatic measurement of sister chromatid exchange frequency. *Journal of Histochemistry and amp; Cytochemistry*, 25(7): 741–753.

Zhuang, J.; Wang, J.; Hoi, S. C. H.; and Lan, X. 2011. Unsupervised Multiple Kernel Learning. In Hsu, C.-N.; and Lee, W. S., eds., *Proceedings of the Asian Conference on Machine Learning*, volume 20 of *Proceedings of Machine Learning Research*, 129–144. South Garden Hotels and Resorts, Taoyuan, Taiwain: PMLR.



\end{document}